\pdfoutput=1

\documentclass[11pt]{article}

\usepackage[acceptedWithA]{tacl2021}
\usepackage{booktabs}
\usepackage{todonotes}
\usepackage{amsmath}
\usepackage{amssymb}
\usepackage{tabularx}
\usepackage{array}

\usepackage{mdframed}
\usepackage{listings}
\usepackage{xcolor}
\usepackage{lipsum}
\usepackage{changepage}
\usepackage{soul}
\usepackage{booktabs}
\usepackage{multirow}
\usepackage{centernot}

\usepackage{CJK,algorithm,algorithmic,amssymb,amsmath,array,epsfig,graphics,float,subfigure,verbatim,epstopdf}

\usepackage{enumitem}
\usepackage{color,soul}
\usepackage{hhline}
\usepackage{multirow}
\usepackage{multicol}

\usepackage{xurl}

\usepackage{tabularx}
\usepackage{bm}
\usepackage{seqsplit}
\usepackage{stmaryrd}
\usepackage{booktabs}

\newmdenv[
    backgroundcolor=lightblue,
    linecolor=darkblue,
    roundcorner=30pt,
    shadow=true,
    shadowsize=2.5pt,
    innertopmargin=10pt,
    innerbottommargin=10pt,
    innerleftmargin=10pt,
    innerrightmargin=10pt,
    linewidth=1.8pt
    width=80pt
]{customframe}

\definecolor{darkgreen}{RGB}{54,128,45}
\definecolor{NavyBlue}{HTML}{1E88E5}
\definecolor{OrangeRed}{HTML}{D81B60}
\definecolor{Gold}{HTML}{dbcd09}

\usepackage{times}
\usepackage{latexsym}
\usepackage[capitalize]{cleveref}

\usepackage[T1]{fontenc}

\usepackage[utf8]{inputenc}

\usepackage{microtype}

\usepackage{inconsolata}

\usepackage{graphicx}

\def\projname{\textsc{SciCo-Radar}}

\title{Inferring Scientific Cross-Document Coreference and Hierarchy with Definition-Augmented Relational Reasoning}

\author{Lior Forer \\ The Hebrew University of Jerusalem \\lior.forer@mail.huji.ac.il\And
   Tom Hope\\The Hebrew University of Jerusalem\\The Allen Institute for AI \\ tomh@allenai.org}

\begin{document}
\maketitle
\begin{abstract}
We address the fundamental task of inferring cross-document coreference and hierarchy in scientific texts, which has important applications in  knowledge graph construction, search, recommendation and discovery. Large Language Models (LLMs) can struggle when faced with many long-tail technical concepts with nuanced variations. We present a novel method which generates context-dependent definitions of concept mentions by retrieving full-text literature, and uses the definitions to enhance detection of cross-document relations. We further generate \emph{relational} definitions, which describe how two concept mentions are related or different, and design an efficient re-ranking approach to address the combinatorial explosion involved in inferring links across papers. In both fine-tuning and in-context learning settings, we achieve large gains in performance on data subsets with high amount of different surfaces forms and ambiguity, that are challenging for models. 
We provide analysis of generated definitions, shedding light on the relational reasoning ability of LLMs over fine-grained scientific concepts.

\end{abstract}

\section{Introduction}
As the volume of scientific papers continues to explode, so does the intricate web of concepts mentioned across papers. Papers in computer science, for example, discuss fine-grained concepts such as algorithms, methods, tasks and problem areas \cite{luan2018multi}. Being able to automatically identify, e.g.,
when papers are referring to the same problem while using very different language (\emph{cross-document coreference}), or that a method used in one paper is part of a more general family of methods (\emph{hierarchy})---could help unlock a wide range of applications, including in scientific knowledge-base construction \cite{mondal2021end, hope2021extracting}, information retrieval \cite{wang2023doris,kang2024improving, wu2024medical}, recommendation systems \cite{mysore2023editable}, systems for paper reading and learning new topics \cite{Murthy2022ACCoRDAM, guo2024personalized}, and even in computational creativity for science \cite{hope2017accelerating,portenoy2022bursting,ji2024scimon,srinivasan2024improving}.
\begin{figure}
    \centering
    \includegraphics[width=\linewidth]{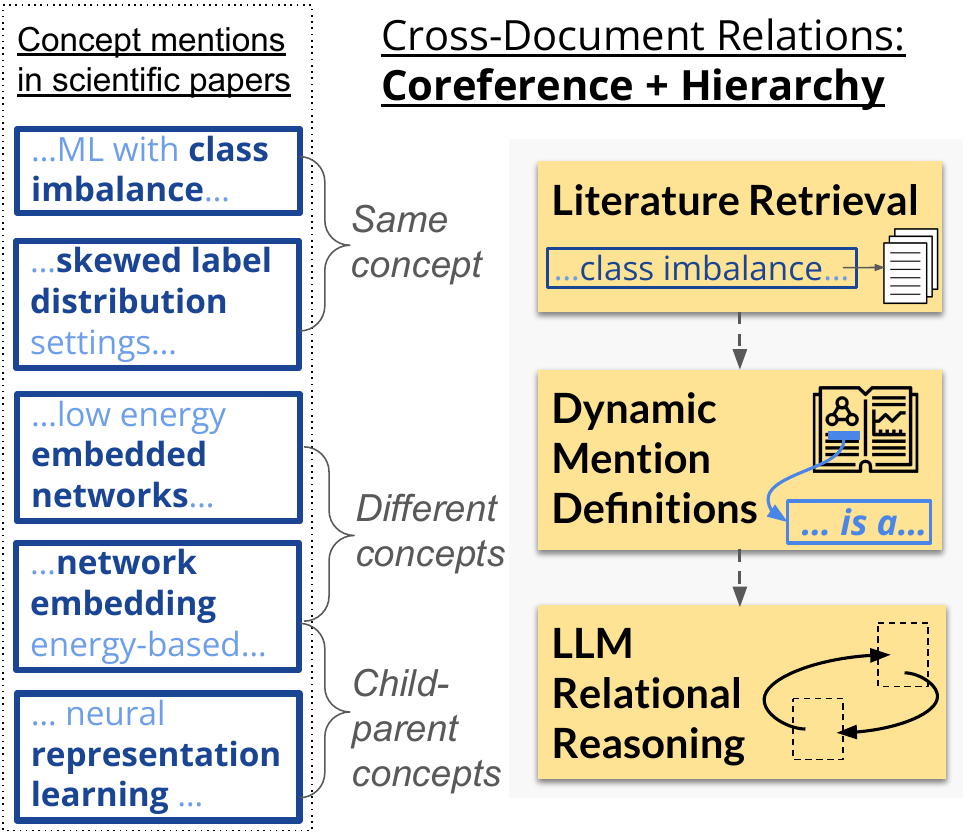}
    \caption{We detect cross-document coreference and hierarchy by augmenting original inputs from papers with context-sensitive definitions and relational reasoning.}
    \label{fig:teaser}
\end{figure}
However, detecting cross-document coreference and hierarchy relations in scientific papers poses significant challenges \cite{cattanscico}. Concepts manifest in long-tail, nuanced variations. Technical language is characterized by ambiguity, different surface forms and fine-grained levels of concept abstractions.  
As an example illustrating several of these challenges, consider \emph{energy-based network embedding}, a specific method under the broader concept of \emph{neural representation learning} despite no surface overlap or lexical cues that may suggest one concept subsumes the other; while \emph{low energy embedded networks} in communications is unrelated, despite strong surface similarity (Figure \ref{fig:teaser}). Surface forms \emph{network}, \emph{energy}, \emph{embedding/embedded} refer to different scientific concepts in different contexts (ambiguity). Another example is \emph{class imbalance} which refers to a \emph{skewed label distribution} (Figure \ref{fig:teaser})---the same underlying scientific concept with different terms.  

We explore the ability of large language models (LLMs) to detect cross-document coreference and hierarchy, i.e., detecting when scientific concept mentions across two papers co-refer (e.g., \emph{network embedding} and \emph{neural representation of graphs}), and when one concept mention is a parent/child of another mention (\emph{network embedding} and the parent concept \emph{neural representation learning}). 

We present a novel method for this challenging setting. 
Given a concept mention in a paper, our approach first generates a concept \emph{definition} (e.g., \emph{network embedding is a neural representation of graphs...}). Definitions are generated by retrieving literature relevant to the mention and synthesizing a concept description that takes into account both global literature information and the local context surrounding the mention. We augment original inputs with the definitions, and train LLMs to make use of them for detecting cross-document coreference and hierarchy. We further introduce \emph{relational} definitions, which explicitly describe how two concept mentions are related or different. To avoid combinatorial explosion when generating relational definitions across all possible pairwise candidates, we create an efficient two-stage re-ranking approach that first generates \emph{singleton} definitions and uses them to score and select candidates for which relational definitions will be computed.   

We train and evaluate our approach on the large and challenging \textsc{SciCo} dataset \cite{cattanscico}, a benchmark for scientific cross-document coreference and hierarchy detection. The examples used above are real cases appearing in \textsc{SciCo}. In both fine-tuning (FT) and in-context learning (ICL) settings, our definition-based approach achieves substantial improvements in detection accuracy, particularly on challenging data subsets marked by a high amount of different surface forms and ambiguity. These hard cases account for 10–20 percent of the test data. 
In the FT setting, we obtain new state-of-the-art results, with singleton and relational definition augmentation leading to large gains in CoNLL F1 over models that do not use definitions, and particularly strong gains in hierarchy detection. In the ICL setting, we evaluate the ability of \texttt{GPT-4o-mini} equipped with the advanced prompt optimizer DSPy \cite{Khattab2023DSPyCD}; we find that when our definition augmentation is added, results substantially increase. 

Finally, we conclude with an extensive analysis of the strengths and limitations of contextualized, relational, and retrieval-based definitions, shedding light on when and how they contribute to performance. This analysis opens up intriguing avenues for future research, particularly in refining the generation and utilization of dynamic concept definitions and in improving the relational reasoning \cite{alexander2016relational} ability of LLMs over nuanced, fine-grained technical concepts.\footnote{Code, models and generated definitions available at \url{https://github.com/TomHopeLab/ScicoRadar}.} 


\section{Problem Setting} 
\label{sec:prob}
\begin{figure*}[h]
    \centering
    \includegraphics[width=\linewidth]{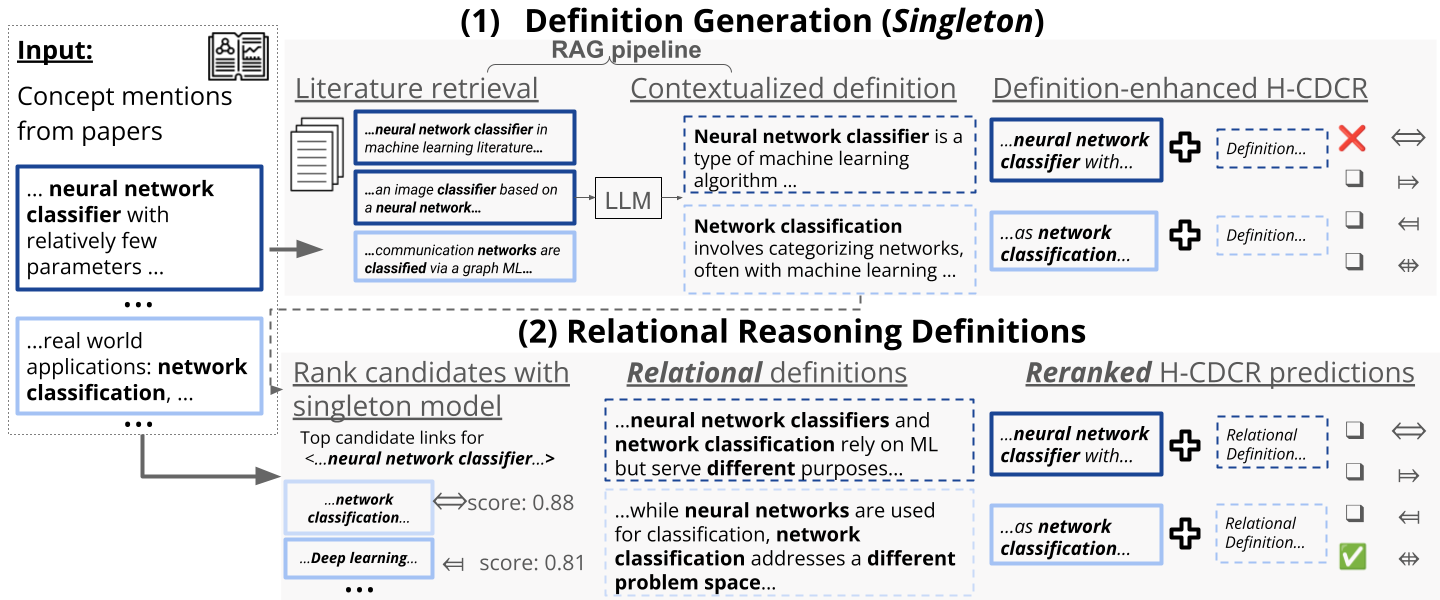}
    \caption{Overview of \projname. We are given as input papers with concept mentions (e.g., methods and tasks). \textbf{(1)} For each mention, we first create \emph{singleton} definitions by retrieving relevant literature and using an LLM to generate context-dependent concept definitions. These definitions are used to augment the original inputs. We use the augmented input to train an LLM to detect cross-document coreference and hierarchy. \textbf{(2)} Using the model trained with singleton definitions, we rank promising top-K candidates, and create for them \emph{relational} definitions that explicitly reason about pairwise concept relationships and further augment the input to enhance detection.}
    \label{fig:main}
\end{figure*}

We focus on the fundamental problem of hierarchical cross document coreference resolution (H-CDCR), first introduced in \citet{cattanscico}. In H-CDCR, our goal is to induce clusters of context-sensitive mentions that co-refer to the same concept, and to deduce a hierarchy among these concept clusters. \citet{cattanscico} created \textsc{SciCo}, a challenging dataset for training and evaluating H-CDCR models focusing on the complex scientific domain; in \textsc{SciCo}, the goal is to infer coreference clusters from sets of mentions extracted from computer science papers, and a hierarchy over the co-reference clusters. For example, as illustrated in Figure \ref{fig:teaser}, the goal would involve detecting that \emph{class imbalance} and \emph{skewed label distribution} belong to the same underlying concept cluster, that \emph{embedded networks} does not belong in the same cluster as \emph{network embedding}, and that \emph{network embedding} belongs to a concept cluster that is a child of the parent concept \emph{representation learning}.

Formally, we are provided with a collection of documents $D$, each $d \in D$ annotated with concept $mentions$. Let $M_d = \{ m_1, m_2, ..., m_n \}$ be the set of mentions in document $d$ and $M$ the set of mentions across all $d \in D$. 
Similar to cross-document co-reference resolution, the first goal is to cluster the mentions in $M$
into disjoint clusters $C = \{C_1, C_2, ..., C_n\}$. Each cluster should contain mentions $\{m|m \in C_i\}$ that
refer to the same underlying concept.
The second goal is to determine a hierarchy over those clusters, specifically a hierarchical relation between cluster pairs $C_i \to C_j$ that reflects on a $referential$ $hierarchy$; a relation $C_i \to C_j$ exists when the concept underlying $C_j$ entails a $reference$ to $C_i$. 
Given documents $D$ and mentions $M$, the goal is to construct clusters $C$ and a hierarchy graph $G_C = (C, E)$ by
learning from a set of N examples $\{(D^k, M^k, G^k_c)\}^N_{k=1}$. 

As we discuss next, we explore learning in both the fine-tuning (FT) and in-context learning (ICL) settings. Our novel approach involves augmenting mentions $M$ with dynamic, contextual \textit{definitions}. We aim to synthesize definitions for specific mentions---and \emph{relational} definitions explicitly describing one or more relations between specific \emph{pairs} of mentions---such that the generated definitions assist models in more accurately inferring the correct cross-document relations.

\section{Methods}
\label{sec:method}
In this section, we present our \projname~(Scientific Concept Induction with Relational Definition Augmented Reasoning) approach for the scientific H-CDCR task. 

\paragraph{Overview} At a high level, our approach consists of several components, as illustrated in Figure \ref{fig:main}. We first retrieve global contexts from scientific papers corresponding to each mention and its local context. Next, we generate a definition for each mention based on the global and local context. We explore two types of definitions. The first type is a \textit{singleton} definition which defines the mention given its context; for example, the definition for \textit{kernel} may correspond to the operating systems or machine learning sense of the term, depending on the context. The second type we explore is a \textit{relational} definition, where a mention is defined explicitly in relation to another mention. 
To avoid combinatorial explosion of generating relational definitions for all possible pairs of mentions and candidates, we further design a candidate re-ranking method inspired by IR research \cite{GUO2020102067}, in which only top-ranked candidates are routed to the relational definition generation module.
Finally, we use the generated definitions to augment mention representations as part of training LLMs on \textsc{SciCo}, in both fine-tuning and in-context learning experiments.
We now provide further details.

\paragraph{Adding definitions to H-CDCR models} Following \citet{cattanscico}, we consider a multiclass classification setting where each mention pair $(m_1, m_2)$ can be assigned into four classes: (1) $m_1$ and $m_2$ co-refer (2) $m_1 \to m_2$ (3) $m_2 \to m_1$ and (0) none. We denote our definition augmentation model as a function $\mathcal{D}(\cdot,\cdot)$, where $\mathcal{D}$ is applied to a pair of candidate mentions $m_1,m_2$ and enriches each mention with a contextual definition. $\mathcal{D}$ may correspond either to singelton definitions, in which case the definitions for $m_1$ and $m_2$ are generated in isolation of each other, or relational definitions in which the definition of $m_1$ depends on $m_2$ and vice versa. Equipped with $\mathcal{D}$ that enriches mention contexts, during training we use a set of mentions $M$ to learn a pairwise scorer $f(*)$ \cite{yang-etal-2022-gpt} that minimizes
\begin{equation}
L = -\frac{1}{N} \sum_{\substack{m_1, m_2 \in \mathcal{M} \\ m_1 \neq m_2}} y \cdot \log(f(\mathcal{D}(m_1, m_2))),
\end{equation}
where $y$ represents one of the four possible classes, $N$ denotes the total number of training pairs, $f(*)$ is a neural language model.

\paragraph{RAG-based definition generation} We use a Retrieval-Augmented Generation (RAG) pipeline for definition generation. Given a specific mention and its context, our method first uses a retriever model to find relevant documents or passages from a large corpus; the retrieved contexts are then fed into a generative model which synthesizes contextualized definitions. Formally, for every mention $m_i \in M$, we denote the passage surrounding $m_i$ as $\phi_{m_i}$ and $\psi(\phi_{m_i})$ is an embedding function that returns a vector representation which we use for retrieving relevant contexts from a large corpus $\mathcal{C}$ of passages obtained from scientific papers.\footnote{We use the corpus of full-text arXiv papers (§\ref{sec:exp}).} 

We embed all passages in $\mathcal{C}$ using $\psi$. Then, for every mention $m_i$ we retrieve a set of passages $\{c_1^i, c_2^i, ..., c_j^i\}$ with maximal dot product similarity to $\psi(\phi_{m_i})$. To reduce noise, we further follow common practice and apply a reranking and filtering step using a re-ranker LM $\xi$ that returns a new re-ranked and  filtered context set $\mathcal{R}_i = \{c_{1^*}, ..., c_{j^*}\} = \xi\left( \{c_1, c_2, \ldots, c_j\} \right)$, where $\mathcal{R}_i$ is an ordered set of contexts for $m_i$, after re-ranking and filtering.
Lastly, we feed $m_i;\phi_{m_i};\mathcal{R}_i$ into 
an LLM that generates contextual definition $d_i$. In Figure \ref{figure:short_unified_prompt}, we show the prompt logic used for generating $d_i$ (see technical details in Appendix \ref{app:definition_generation}).

\paragraph{Relational definitions} We introduce another method for generating definitions, which we name as \emph{relational}. Relational definitions are designed to describe mentions in connection to another mention, thereby providing a more direct, explicit understanding of how two mentions may be related to each other. To create relational definitions, we feed into an LLM the \emph{pairwise} contexts of two mentions, $m_i;\phi_{m_i};\mathcal{R}_i$ and $m_j;\phi_{m_j};\mathcal{R}_j$, with a custom prompt (Figure \ref{figure:short_unified_prompt}) for defining $m_i$ in relation to $m_j$ or vice versa. Note that relational definitions are asymmetric as they are centered around an ``anchor'' mention (see examples in Figure \ref{fig:main} and Table \ref{tab:example_definitions}).

However, generating relational definitions for all pairs of mentions quickly explodes; for instance, in the \textsc{SciCo} test set this would entail 564,352 unique pairs for which we would need to generate definitions, where each mention contains multiple contexts from papers as well as the surrounding paragraph. On the other hand, singleton definitions may be computed offline for individual mentions and stored and retrieved efficiently.
Thus, in order to make the relational definition generation process feasible,  we create a two-stage method in which the first stage consists of using singleton definitions to efficiently filter the space of possible candidates, and in the second stage we create relational definitions only for a small subset of candidates. This approach broadly takes after the common retrieve-and-rerank pipeline often used in information retrieval systems \cite{linpretrained} and also cascade approaches to link prediction \cite{safavicascader}.

More formally, for each mention $m_i \in M$ and each candidate $m_j$ we use singleton definition augmentation to compute likelihood scores $f(\mathcal{D}(m_i, m_j))$ (each prediction of the model is a logit of size 4). We first filter out pairs for which the model predicted \textit{none} (no relationship). Then, for each $f(\mathcal{D}(m_i, m_j))$ we compute the softmax probabilities and rank the predictions based on the model's confidence in its assigned relation. We set a threshold value $\theta$ to select the top 25\% pairs as a tradeoff between accuracy and computational efficiency; we compare different values of this threshold on the \emph{Hard-10} subset in Appendix \ref{app:ablation}). For each mention $m_i$, we generate relational definitions solely for the top pairs of mentions for which $f(\mathcal{D}(m_i, m_j)) \geq \theta$. 

\paragraph{\texttt{GPT-4o} based definition generation} In addition to the RAG-based approach, we explore the use of \texttt{GPT-4o} \cite{gpt4o} for generating definitions. This method leverages the model's extensive parametric knowledge without incorporating external context retrieval. Specifically, for each mention $m_i$, we provide \texttt{GPT-4o} with the passage surrounding $m_i$ denoted as $\phi_{m_i}$. Unlike the RAG pipeline, this approach does not utilize additional context from external sources. While this method simplifies the definition generation process, it may reduce the quality of generated definitions, as we explore further in Sections \ref{sub-sec:results} and \ref{sec:qual}; it also substantially increases costs.

\paragraph{LLM training} Finally, given definitions we integrate them into LLMs for training. The current state of the art reported by \citet{cattanscico} was based on fine-tuning a \textsc{Longformer} model \cite{Beltagy2020Longformer}. We fine-tune recent state of the art LLM, and also experiment with both few-shot prompting and the more advanced DSPy \cite{Khattab2023DSPyCD} prompt optimization framework (see more details in Appendix \ref{app:dspy}).

\section{Experiments}


\label{sec:exp}
In this section we describe our experimental setup and results. We first outline the implementation details (§\ref{sec:implementation_details}), and then present the main results (§\ref{sub-sec:results}), which introduce the baselines and report performance on both the hardest subsets and the overall dataset.

\subsection{Implementation Details}
\label{sec:implementation_details}
\paragraph{Finetuning with definitions} 

We train models both with the inclusion of definitions and without them. As base models, we report results for fine-tuning
\textsc{Mistral-7B}\footnote{To check generalizablity of our results beyond Mistral, we also conduct a brief ablation experiment with another recent open source LLM \cite{abdin2024phi}.  We observe similar overall trends. See Tables \ref{tab:Hierarchy_all}--\ref{tab:co-ref_hard_10} in Appendix.} \cite{Jiang2023Mistral7}, and \textsc{Longformer} \cite{Beltagy2020Longformer} which was used originally in \citet{cattanscico}. 
We augment the original data to inject a definition for every mention. For
\textsc{Mistral-7B} we construct a prompt (Figure \ref{figure:finetuning_mistral_classification}) following the ChatML\footnote{ \url{github.com/openai/openai-python/blob/release-v0.28.0/chatml.md}} 
format. For Longformer the input is processed as follows. For each mention $m1, m2$ and its corresponding paragraph, mention markers <m> and </m> surrounding the mention are used for obtaining a mention representation. We insert additional <def> and </def> markers surrounding each definition $d_{m_1}, d_{m_2}$. Finally we concatenate all the above with the separation tokens <s> and </s> separating each mention (See prompt in Figure \ref{figure:Longformer_format}).



We fine-tuned the \textsc{Mistral-7B}
model from the HuggingFace \cite{Wolf2019HuggingFacesTS} library by adding a classification head, replacing the original head with a task-specific classification layer.\footnote{\url{huggingface.co/transformers/v3.0.2/model_doc/auto.html\#automodelforsequenceclassification}} The classification head is a simple feed-forward layer with a softmax activation function (see Appendix \ref{app:hardware_lora} for hyper parameters, hardware and implementation details). We also attempted an alternative generative fine-tuning approach with the same setup, where the model was tasked with generating the correct relationship for each pair. However, this approach led to worse results in our experiments. 

\paragraph{ICL} In our experiments with few-shot in-context learning (ICL) we use two types of prompts: one incorporating definitions and the other without them. We include one example per class (four total; DSPy further optimizes in-context examples). The prompt with definitions provides additional context for each example (see prompt in Figure~\ref{figure:few_shots_with_def_format}), while the prompt without definitions relies solely on the examples (Figure~\ref{figure:few_shots_no_def_format}). To account for variability in results due to few-shot selection, we report average results for 5 runs for each configuration, each time sampling a different set of few-shot examples and corresponding definitions (within each run, the same ICL examples are used for both with/without definition settings). We structure the output to extract classifications and reasoning automatically.\footnote{\url{dspy-docs.vercel.app/docs/deep-dive/signature/executing-signatures\#inspecting-output}.} 

Due to the combinatorial nature of our task with cross-document links across a potentially huge candidate space, we focus ICL evaluation on a strong yet comparatively cheap model, \texttt{GPT-4o-mini}, and applied only to a subset of the test set (§\ref{sub-sec:results}). Our ICL experiments cost roughly \$500, while a stronger \texttt{GPT-4o} model would have incurred a cost of over \$16000 which is beyond our academic budget. Concurrent results \cite{lior2024seam} on \textsc{SciCo} with various open-source \textsc{Mistral} models in the few-shot setting led to very poor results in comparison to our results with \texttt{GPT-4o-mini} even without our definition-augmentation approach.


\begin{figure*}
    \centering
    \includegraphics[width=\linewidth]{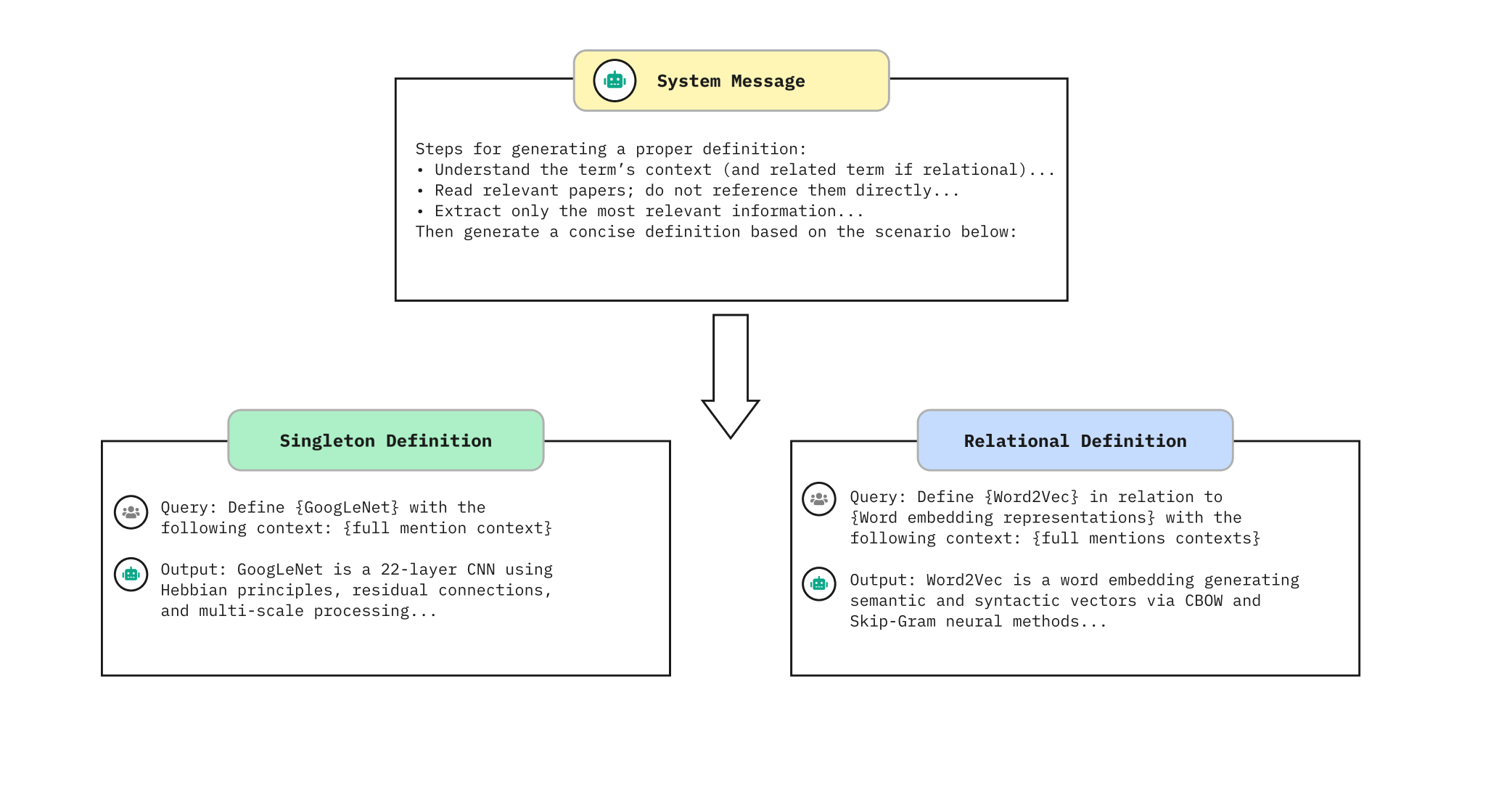}
    \caption{Logic and structure of the prompt used to generate singleton and relational definitions.}
    \label{figure:short_unified_prompt}
\end{figure*}

\label{paragraph: evaluation_metrics}
\paragraph{Evaluation Metrics}
For co-reference evaluation, we adopt the standard metrics MUC, B$^3$, CEAFe, and LEA, and report the widely used \emph{CoNLL F1}, defined as the average of the MUC, B³, and CEAFe F1 scores, as our primary measure.
To evaluate the hierarchy, we use the following evaluation methods created in \citet{cattanscico}:
\begin{itemize}
    \item \textbf{Cluster-Level Hierarchy Score} checks whether each predicted parent--child link between mention clusters aligns with \textit{at least one} mention pair in the gold hierarchy. This design avoids ``double penalizing'' minor co-reference mismatches, For example, if the system predicts a parent–child relationship between clusters [Information Extraction] → [Definition Extraction, Pattern-based Extraction], and at least one pair of mentions (e.g., ``Information Extraction'' → ``Definition Extraction'') exists in the gold hierarchy, this counts as correct. We also consider a stricter version (50\% Hierarchy Score) that requires \emph{at least half} of the mentions in the predicted cluster pair to align with the gold parent--child link.
    \item \textbf{Path-Distance Score} compares distances between mentions (in terms of the number of edges separating them) in the predicted hierarchy vs.\ the gold. It provides partial credit when the distances are close, even if not perfectly matched. For instance, consider the mentions ``Information Extraction'' and ``Pattern-based Definition Extraction'': if they are separated by two edges in the gold hierarchy but only one edge in the predicted hierarchy, the metric gives partial credit proportional to the closeness of these path lengths.
\end{itemize}
We refer the reader to \citet{cattanscico} for full technical details on these hierarchy metrics.

\begin{table*}[t]
\centering
\footnotesize
\renewcommand{\arraystretch}{1.15}
\begin{tabular}{@{}lcccc@{}} 
\toprule
\textbf{Model} & \textbf{Coreference} & \multicolumn{2}{c}{\textbf{Hierarchy}} & \textbf{Path} \\
               & \textbf{CoNLL F1} & \textbf{F1} & \textbf{F1-50\%} & \textbf{Ratio} \\
\midrule
\textit{\textsc{SciCo-Longformer}}\cite{cattanscico} & 64.34 & 45.46 & 28.07 & 34.26 \\
\midrule
\textit{\textsc{GPT 4o-mini}\textsubscript{}} & 51.28 & 40.22 & 16.95 & 31.04 \\
\textit{\textsc{GPT 4o-mini}\textsubscript{Singleton}} & 54.91 & 43.33 & 20.13 & 32.56 \\
\textit{\textsc{SciCo-Mistral-7B}} & 69.68 & 50.25 & 37.36 & 49.55 \\
\textit{\textsc{\projname}\textsubscript{Singleton}} & 70.30 & 54.70 & 43.40 & 50.00 \\
\textit{\textsc{\projname}\textsubscript{Relational}} & \textbf{71.90} & \textbf{56.98} & \textbf{45.93} & \textbf{53.85} \\

\bottomrule
\end{tabular}
\caption{Results on top 10\% hardest topics (\emph{Hard-10} subset). \texttt{GPT-4o-mini} uses few-shot examples (one per class). \texttt{GPT-4o-mini}\textsubscript{\emph{Singleton}} adds our singleton definitions, which increase performance by several points. Fine-tuning \textsc{Mistral-7B} on \textsc{SciCo} leads to a large improvement. \projname~refers to \textsc{Mistral-7B} fine-tuned on  \textsc{SciCo} along with our definition augmentation approach. \projname~ leads to a further large boost in results, particularly in terms of hierarchy detection. \projname~with relational definitions achieves the best results. }
\label{tab:combined_hard_10}
\end{table*}

\subsection{Results}
\label{sub-sec:results}
We next present the experimental results. We begin by establishing baselines and evaluating on the hardest subsets of the dataset, where the task is most challenging and the benefits of our definition based approach has the most impact. We then assess performance on the full dataset to measure overall improvements, and conclude with an ablation study that examines the specific contribution of different types of definitions.

\begin{table*}[t]
\centering
\footnotesize
\renewcommand{\arraystretch}{1.2}
\setlength{\tabcolsep}{6pt}
\begin{tabularx}{\textwidth}{
    l
    *{3}{>{\centering\arraybackslash}X}
    !{\hspace{10pt}\vrule width 1pt\hspace{10pt}}
    *{3}{>{\centering\arraybackslash}X}
}
\toprule
\textbf{Metric} &
\multicolumn{3}{c}{\textsc{SciCo-Longformer} (Baseline)} &
\multicolumn{3}{c}{\textsc{SciCo-Longformer}\textsubscript{Singleton def}} \\
\midrule
& \textbf{All} & \textbf{\emph{Hard-20}} & \textbf{\emph{Hard-10}}
& \textbf{All} & \textbf{\emph{Hard-20}} & \textbf{\emph{Hard-10}} \\
\midrule
\textbf{CoNLL F1}          & 76.46 & 68.18 & 64.34 & 78.18 & 71.10 & 65.28 \\
\textbf{Hierarchy F1}      & 46.09 & 46.44 & 45.46 & 47.43 & 49.28 & 46.21 \\
\textbf{Hierarchy F1 50\%} & 35.14 & 29.61 & 28.07 & 38.52 & 33.77 & 29.52 \\
\textbf{Path Ratio}        & 44.91 & 36.21 & 34.26 & 48.81 & 44.07 & 35.36 \\
\bottomrule
\end{tabularx}
\caption{Results for \textsc{SciCo-Longformer} (Baseline) and \textsc{SciCo-Longformer}\textsubscript{Singleton def} across the full \textsc{SCICO} test set and two harder subsets (\emph{Hard-20}, \emph{Hard-10}). Metrics (§\ref{paragraph: evaluation_metrics}) shown are CoNLL F1, cluster-level hierarchy F1 as well as 50\% and path ratio.}
\label{tab:longformer_all_data}
\end{table*}

\paragraph{Evaluation focus}
In addition to reporting overall performance (see Table \ref{tab:combined_all}), we focus on the ``hard'' data subsets provided by \citet{cattanscico}, comprised of the 10\% (\emph{Hard-10}) or 20\% (\emph{Hard-20}) topics (collections of candidate mentions) with lowest scores as ranked according to CoNLL F1 by using Levenshtein distance in agglomerative clustering as a baseline. The subsets are particularly challenging, with a high amount of different surface forms and ambiguity, and are thus of most interest; e.g., the best model reported by \citet{cattanscico}, achieved only 64.34 F1 for coreference CoNLL F1 on the \emph{Hard-10} subset, while achieving 76.46 F1 on the full test set.

\begin{table*}
\centering
\begin{tabular}{lcccc} 
\toprule
\textbf{Model} & \textbf{Coreference} & \multicolumn{2}{c}{\textbf{Hierarchy}} & \textbf{Path} \\
               & \textbf{CoNLL F1} & \textbf{F1} & \textbf{F1-50\%} & \textbf{Ratio} \\
\midrule
\textit{\textsc{SciCo-Longformer}} (Baseline) & 76.46 & 46.09 & 35.14 & 44.91 \\
\midrule
\textit{\textsc{SciCo-Mistral-7B}} & 81.82 & 56.84 & 49.70 & 57.21 \\
\textit{\textsc{\projname}\textsubscript{Singleton}} & 81.95 & 57.33 & 50.58 & 57.27 \\
\textit{\textsc{\projname}\textsubscript{Relational def}} & \textbf{82.07} & \textbf{58.03} & \textbf{50.72} & \textbf{57.60} \\
\bottomrule
\end{tabular}
\caption{Results on the full \textsc{SciCo} test set. Fine-tuning Mistral-7B substantially outperforms the Longformer baseline, and adding singleton or relational definitions gives further improvements, with relational definitions achieving the best overall scores.}
\label{tab:combined_all}
\end{table*}

\paragraph{Baselines} As a primary baseline, we use the model introduced in the original \textsc{SciCo} paper \citet{cattanscico}, which defines the standard evaluation setting for this task. In addition, we include our own fine-tuned models that do not use definitions, in order to isolate the contribution of definition based augmentation. We also report results for in-context learning (\textsc{ICL}) with GPT-4o-mini, to establish a baseline for definition augmentation without finetuning. While SEAM \citet{lior2024seam} also reports results on \textsc{SciCo} and uses more recent models, its performance falls well below that of the original \textsc{SciCo} baseline.

\paragraph{Evaluation summary}
We report our results in Table \ref{tab:combined_hard_10} (\emph{Hard-10}), Table \ref{tab:longformer_all_data} (Longformer all data) and Table \ref{tab:combined_all} (all data) from the main paper, along with the full Tables [\ref{tab:Hierarchy_all}, \ref{tab:Hierarchy_hard20}, \ref{tab:Hierarchy_hard10}, \ref{tab:co-ref_all}, \ref{tab:co-ref_hard_20}, \ref{tab:co-ref_hard_10}] in the Appendix (results in all subsets reflect similar model trends). ICL few-shot results are averaged across 5 runs as described earlier. We first observe that, not very surprisingly, fine-tuning of recent LLMs leads to large gains over the smaller \textsc{Longformer} model used originally in \citet{cattanscico}. Fine-tuning also achieve considerable gains over concurrent results for \textsc{SciCo} CDCR in the few-shot setting with larger variants from the \textsc{Mistral} family (see Table 3 in \citet{lior2024seam}), highlighting the value of fine-tuning in this task. More interestingly, our contextualized definitions led to strong improvements, for both LLM fine tuning and ICL. Our \projname\textsubscript{Relational}~model, which consists of fine-tuning \textsc{Mistral-7B} on \textsc{SciCo} augmented by relational definitions, achieved the highest results in every evaluation category, as exemplified in Table \ref{tab:combined_hard_10}. Generally, relational definitions outperformed singleton definitions, by carrying more nuanced information about relationships that the model might miss with the more generic singleton definitions\footnote{Paired bootstrap tests (Appendix~\ref{app:significance}) confirm statistically reliable gains for hierarchy metrics across both hard subsets and for CoNLL F1 on \emph{Hard-10}, with the remaining metrics reflecting positive upward trends.} (see Section \ref{sec:qual} for more analysis). ICL with larger LLMs \cite{gpt4o}, including prompt optimization with DSPy, underperformed smaller fine-tuned models in our experiments, though future work may explore this further with more models and in-context examples.

As seen in Table \ref{tab:longformer_all_data}, augmenting  \textsc{Longformer} with our contextual definitions also resulted in substantial enhancements in both hierarchy and coreference scores, by up to 3.38 and 1.72 respectively. Even more substantial improvements were observed within the \emph{Hard-20} subsets, particularly in hierarchy and hierarchy 50\% scores, achieving improvements of 2.84 and 4.16. With
\textsc{Mistral-7B}, incorporating definitions consistently improved all metrics. As seen In Table \ref{tab:combined_hard_10}, incorporating definitions notably boosted \textsc{Mistral}'s hierarchy F1 and F1-50\% scores by up to 4.5 and 6.1 using singleton definitions, and by 6.78 and 8.63 with relational definitions.

We also report additional coreference and hierarchy precision-recall metrics in Tables \ref{tab:Hierarchy_all}, \ref{tab:Hierarchy_hard20} and \ref{tab:co-ref_all}, \ref{tab:co-ref_hard_20} in the Appendix. For coreference, the F1 gains with relational definitions come from a substantial boost in recall with only a relatively small dip in precision. For hierarchy, we see a similar trend where both singleton and relational definitions improve recall while maintaining stable precision. 

Finally, to assess the stability of ICL results with random few-shot examples or examples optimized by random search in DSPy, we report average results for five runs (Table \ref{tab:gpt-5run}).\footnote{The \textsc{SciCo} dataset contains 69,682 mention pairs in the \emph{Hard-10} subset, along with extensive context and definitions per mention; running our ICL experiments on this subset cost \$500. The full test set contains 564,352 mention pairs.} We find that the effects are overall stable, with definitions helping boost results across both the few-shot and DSPy experiments;\footnote{We also experimented with several runs of fine-tuning, finding negligible variance across seeds.} interestingly, we do not observe gains from using DSPy in our experiments.

\begin{table*}
\centering
\footnotesize
\renewcommand{\arraystretch}{1.15}
\begin{tabular}{@{}lcccc@{}} 
\toprule
\textbf{Model} & \textbf{Coreference} & \multicolumn{2}{c}{\textbf{Hierarchy}} & \textbf{Path} \\
               & \textbf{CoNLL F1} & \textbf{F1} & \textbf{F1-50\%} & \textbf{Ratio} \\
\midrule
\textit{\textsc{GPT 4o-mini}\textsubscript{}} & 51.28 ± 2.21 & 40.22 ± 1.70 & 16.95 ± 1.62 & 31.04 ± 2.23 \\
\textit{\textsc{GPT 4o-mini}\textsubscript{Singleton def}} & 54.91 ± 0.61 & 43.33 ± 0.96 & 20.13 ± 0.61 & 32.56 ± 1.07 \\
\textit{\textsc{GPT 4o-mini}\textsubscript{DSPy}} & 50.71 ± 0.66 & 42.23 ± 0.94 & 14.74 ± 1.36 & 27.41 ± 1.29 \\
\textit{\textsc{GPT 4o-mini}\textsubscript{DSPy Singleton def}} & 54.90 ± 0.52 & 43.72 ± 1.16 & 19.69 ± 1.97 & 30.70 ± 0.56 \\
\bottomrule
\end{tabular}
\caption{\texttt{GPT-4o-mini} ablation study. Co-reference and hierarchy scores with different configurations, average results and SE (five runs) on the \emph{Hard-10} subset. We see the positive effects of definitions with and without DSPy optimization.  We focus on \texttt{GPT-4o-mini} due to its relatively affordable cost (\$500 to produce this table) and speed, important in our setting of inferring cross-document links in large-scale literature.}
\label{tab:gpt-5run}
\end{table*}

\label{sub-overall-vs-hard}
\paragraph{Overall vs. Hard Subsets}
Table~\ref{tab:combined_all} shows only modest overall improvements on the full \textsc{SciCo} test set, whereas Table~\ref{tab:combined_hard_10} reveals much larger gains on the \emph{Hard-10} and \emph{Hard-20} subsets. This is expected, since many mentions can be resolved by simple lexical overlap or the model's prior knowledge, reducing the benefits of adding definitions. By contrast, the \emph{Hard-10}/\emph{Hard-20} cases feature ambiguous concepts and more variations of different surface forms, precisely where retrieval-based definitions can clarify those subtle distinctions or shared properties, thus offering substantial improvements. Additionally, these challenging cases make up only 10--20\% of the data, so the strong  gains do not dramatically affect the overall average. Nonetheless, linking these rare or nuanced mentions accurately is crucial for downstream scientific tasks, for example, knowledge graph construction or document retrieval, where mislinking can lead to incorrect representations of fine-grained concepts. A similar phenomenon was observed by \citet{ravi-etal-2023-happens} in event coreference, where the majority of mention pairs were straightforwardly matched by a model’s surface or parametric knowledge. Substantial benefits from injecting external knowledge was reported primarily for the smaller more challenging fraction of mentions, in line with our findings on the \emph{Hard-10}/\emph{Hard-20} subsets. Notably, in table \ref{tab:longformer_all_data}, we observe that the gains from adding definitions for the \textsc{Longformer} model are larger on the full and the \emph{Hard-20} test sets compared to the \emph{Hard-10} subset, which contrasts with the trends we observed for stronger models. We hypothesize that \textsc{Longformer} struggles more with the inherently ambiguous and challenging examples in the Hard 10 subset, limiting its ability to fully exploit the additional context provided by definitions. Consequently, while definition augmentation still improves performance, the relative gains are smaller on the hardest cases compared to simpler ones.

\begin{table*}[h!]
\resizebox{0.99\textwidth}{!}{%
\begin{tabular}{@{}>{\raggedright\arraybackslash}p{6cm} p{6cm} p{7cm}@{}}
\toprule
\textbf{Context and Prediction} & \textbf{Singleton Def} & \textbf{Relational Def} \\ 
\midrule
\textbf{1}: ... WAN expansion regardless of network structures and to enhance the \textbf{\emph{network cyber-security}}...  & \textbf{1}: Protecting networks and information from unauthorized access ... firewalls, antivirus software, encryption... & \textbf{1}: Protecting networks ... firewalls, encryption ... In contrast to cyber-safety awareness individual behavior and education ... \\ \\
 \textbf{2}: ... situation concerning \textbf{\emph{cyber-safety awareness}} in schools and has adopted a short-term approach towards cyber-safety ...
 \newline\textcolor{darkgreen}{\textbf{Ground Truth: $\centernot\Longleftrightarrow$}}\newline\textcolor{red}{\textbf{No Def Model: $\Longleftrightarrow$}} & \textbf{2}: Recognition of potential online threats ... educating individuals ... about protecting personal information ... \newline\textcolor{darkgreen}{\textbf{Correct Prediction: $\centernot\Longleftrightarrow$}} & \textbf{2}: Recognition of potential online threats ... Unlike network cyber-security, which protects systems, cyber-safety awareness seeks to protect and inform users ... \newline\textcolor{darkgreen}{\textbf{Correct Prediction: $\centernot\Longleftrightarrow$}}\\
 
\midrule

\textbf{1}: ... they can use a technology for \textbf{\emph{online marketing}} using e-commerce ...  & \textbf{1:} Using digital channels to promote products or services... web advertising, e-commerce, targeted promotions based on customer data... & \textbf{1}: Using digital channels to promote products or services... closely related to targeted ads, which are a key component of online marketing strategies... \\ \\

 \textbf{2}: ... data gathering for purposes of personalization, \textbf{\emph{targeted ads}}... 
 \newline\textcolor{darkgreen}{\textbf{Ground Truth: $\Longrightarrow$}} \newline\textcolor{red}{\textbf{No Def Model: $\centernot\Longleftrightarrow$}} & \textbf{2:} Online advertisements  tailored to individuals or groups... \newline\textcolor{darkgreen}{\textbf{Correct Prediction: $\Longrightarrow$}} &\textbf{2}: A form of online marketing ... often employs sophisticated user profiling methods ... \newline\textcolor{darkgreen}{\textbf{Correct Prediction: $\Longrightarrow$}}\\

\midrule
\textbf{1}: ... These findings provide new possibilities of systematic F0 control for \textbf{\emph{conversational speech synthesis}} ...  & \textbf{1}: The generation of speech that mimics natural, everyday conversation, often utilizing machine learning techniques to control ... & \textbf{1}: the generation of speech that mimics natural, everyday conversation, often utilizing machine learning techniques ... It can be distinguished from traditional speech synthesis ... \\ \\
 \textbf{2}: ...  SND and diarization can then be used ... achieving modern performance in \textbf{\emph{conversational speech processing tasks}} ...
 \newline\textcolor{darkgreen}{\textbf{Ground Truth: $\Longleftarrow$}}\newline\textcolor{red}{\textbf{No Def Model: $\centernot\Longleftrightarrow$}} & \textbf{2}: A series of computational procedures applied to spoken language during social interactions ... encompass Speech Separation, Speaker Diarization, Automatic Speech Recognition ... \newline\textcolor{darkgreen}{\textbf{Correct Prediction: $\Longleftarrow$}} & \textbf{2}: Operations involved in analyzing and understanding spoken language, including Speech Separation, Diarization, Automatic Speech Recognition ...  a precursor to applications such as conversational speech synthesis ... \newline\textcolor{darkgreen}{\textbf{Correct Prediction: $\Longleftarrow$}}\\
 
\bottomrule
\end{tabular}
}
\caption{Examples where the model without definitions misidentifies the true relation. The singleton and relational definitions add important context enabling the model to accurately determine the correct relation. $\centernot\Longleftrightarrow$ - no relationship, $\Longleftrightarrow$ - co-reference, $\Longrightarrow$ - 1 $\rightarrow$ 2 parent-child relationship and $\Longleftarrow$ - 2 $\rightarrow$ 1 parent-child relation.}
\label{tab:example_definitions}
\end{table*}

\paragraph{\texttt{GPT-4o} definitions} As an additional baseline and ablation, we use the powerful and more costly \texttt{GPT-4o} to generate definitions by providing the model with the mention and its surrounding context but no additional retrieval, using only the model's vast knowledge embedded in its parameter space \cite{Ye2023ACC,Bubeck2023SparksOA}. We find that after training \textsc{Mistral-7B} on these definitions, a big gap in performance emerged in favor of retrieval-based definitions in terms of hierarchy predictions (Table \ref{tab:defablation_hard_10}), along with a smaller but noticeable advantage in coreference scores, too (Table \ref{tab:co-ref_hard_10}). Upon reviewing the generated definitions, we observed that \texttt{GPT-4o} often generated definitions that were more generic and tended to be more fixated on irrelevant context details, while retrieval-based definitions naturally assisted the definition generation to be both more detailed and specific while zooming out and abstracting-away irrelevant context. For example, in Table \ref{tab:example_retrieval_definitions}, the GPT-4o definition of ``representation learning'' is too focused on a specific application involving embodied agents and cognitive phenomena, while the singleton definition abstracts away from these details and instead highlights the broader machine learning context, the kind of generalization needed to resolve this hierarchical relationship (Table \ref{tab:example_retrieval_definitions} in the Appendix; §\ref{sec:qual} for more analysis).

\begin{table*}[t]
\resizebox{0.99\textwidth}{!}{%
\begin{tabular}{@{}>{\raggedright\arraybackslash}p{6cm} p{6cm} p{7cm}@{}}
\toprule
\textbf{Context and Prediction} & \textbf{Singleton Def} & \textbf{Relational Def} \\ 
\midrule

\textbf{1}: ... Then the RNN language models predict probability distribution by the following equation ... is a \textbf{\emph{word embedding matrix network}} ...  & \textbf{1}: A matrix in which each row corresponds to a word in a given vocabulary ... used in natural language processing models to project discrete word tokens into a continuous vector space ...  & \textbf{1}: A matrix where each row corresponds to a word in a given vocabulary ... This matrix is used to project discrete word tokens into distributed representations of words, which are dense vectors ... \\ \\

\textbf{2}: ... RNN can utilize \textbf{\emph{distributed representations of words}} by first converting ...  
\newline\textcolor{darkgreen}{\textbf{Ground Truth: $\Longleftrightarrow$}} 
\newline\textcolor{red}{\textbf{No Def Prediction: $\centernot\Longleftrightarrow$}} & \textbf{2}: The way words are represented as vectors in computational models ... These vectors capture semantic and syntactic properties of words ...  
\newline\textcolor{red}{\textbf{Wrong Prediction: $\centernot\Longleftrightarrow$}} & \textbf{2}: The way words are represented as vectors ... In the context of RNNs, these distributed representations are utilized to convert tokens into vectors, forming a matrix that captures ... 
\newline\textcolor{darkgreen}{\textbf{Correct Prediction: $\Longleftrightarrow$}}\\
\midrule
\textbf{1}: ... the \textbf{\emph{simulated-annealing-based approach}} is proposed to construct a better timing-constrained ...  & \textbf{1}: A computational technique used to optimize a problem by iteratively improving an initial solution through a series of modifications ...  & \textbf{1}: A type of meta-heuristic approach that utilizes the principles of annealing used to optimize a problem by iteratively improving an initial solution through ... \\ \\
\textbf{2}: ... results of numerous implementations of well-known \textbf{\emph{meta-heuristic approaches}} ...  
\newline\textcolor{darkgreen}{\textbf{Ground Truth: $\Longleftarrow$}} 
\newline\textcolor{red}{\textbf{No Def Prediction: $\centernot\Longleftrightarrow$}} & \textbf{2}: A category of optimization algorithms that include Evolutionary Algorithms, Genetic Algorithms, and Swarm Intelligence techniques among others ... 
\newline\textcolor{red}{\textbf{Wrong Prediction: $\centernot\Longleftrightarrow$}} & \textbf{2}: A class of high-level, general-purpose optimization algorithms that ... These approaches include Evolutionary Algorithms, Simulated Annealing, Swarm Intelligence, and others ... 
\newline\textcolor{darkgreen}{\textbf{Correct Prediction: $\Longleftarrow$}}\\
\midrule
\textbf{1}: ... \textbf{\emph{neural network classifier}} employs relatively few parameters compared to other deep learning ...  & \textbf{1}: A type of machine learning algorithm that uses artificial neural networks to categorize data into different classes ...  & \textbf{1}: A type of machine learning algorithm that uses artificial neural networks to classify ... Although both neural network classifiers and network classification rely on machine learning techniques, they serve different purposes ... \\ \\

\textbf{2}: ... real world applications such as \textbf{\emph{network classification}}, and anomaly detection ...  
\newline\textcolor{darkgreen}{\textbf{Ground Truth: $\centernot\Longleftrightarrow$}} 
\newline\textcolor{red}{\textbf{No Def Prediction: $\Longleftrightarrow$}} & \textbf{2}: A process that involves categorizing networks or graphs based on their structure, often utilizing machine learning algorithms ... 
\newline\textcolor{red}{\textbf{Wrong Prediction: $\Longleftrightarrow$}} & \textbf{2}: The process of categorizing networks or their components based on ... While neural networks can be used for classification tasks, network classification addresses a different problem space, focusing on ... 
\newline\textcolor{darkgreen}{\textbf{Correct Prediction: $\centernot\Longleftrightarrow$}}\\

\bottomrule
\end{tabular}
}
\caption{Analysis of relational definitions. Relational definitions can help clarify mention relationships in cases where singleton definitions are unable. $\centernot\Longleftrightarrow$ - no relationship, $\Longleftrightarrow$ - co-reference, $\Longrightarrow$ - 1 $\rightarrow$ 2 parent-child relationship and $\Longleftarrow$ - 2 $\rightarrow$ 1 parent-child relationship.}
\label{tab:example_relational_definitions}
\end{table*}

\section{Qualitative Analysis}
\label{sec:qual}

In this section, we explore the impact of definitions on model performance by presenting qualitative examples and analyzing their effects. We also perform error analysis, revealing different sources of error which may inform future improvements. Our focus is on understanding how different types of definitions can influence the output of the model, specifically in cases where the base model struggles. We use our best-performing fine-tuned \projname~model for these analyses. 

\subsection{When Definitions Generally Help} We begin with providing qualitative evidence to support the fundamental hypothesis in this paper, showing that incorporating dynamic definitions substantially improves model performance over the non-definition approach. 
We discuss two main ways in which we qualitatively observe definitions broadly help, before diving deeper into specific types of definitions and their various trade-offs.

\paragraph{Lexical ambiguity, different surface forms, and fine-grained hierarchy} Lexical ambiguity, different surface forms, and fine-grained concept hierarchies are at the heart of the challenge for the H-CDCR task in complex technical domains. Our findings confirm our hypothesis that definitions provide useful context that helps models understand lexical and semantic nuances for resolving the true relationships of mentions, or lack there of, by detecting shared meanings and aspects that are not obvious from the text alone. For instance (see also Table \ref{tab:example_definitions}, row 2), the definition of \emph{online marketing operation} describes it as encompassing various digital strategies, including \emph{targeted ads}, clarifying that the former subsumes the latter. Similarly, defining \emph{targeted ads} as a specific type of online advertisement tailored to specific demographics makes the hierarchical relationship clear. As another example, demonstrating lexical ambiguity (see also Table \ref{tab:example_definitions}, row 1), the definitions for \emph{network cyber-security} and \emph{cyber-safety awareness} make it clear that one mention focuses on technical measures to secure network infrastructure, while the other emphasizes educating individuals about online safety practices. These definitions help the model understand that, despite surface similarities, the mentions do not belong to the same conceptual cluster or share a hierarchical link.

\paragraph{Context lacking information \& context interference} We observe that, unsurprisingly, vague or ambiguous contexts can make it hard to clarify whether two different mentions are co-referential, hierarchical, or unrelated.
Additionally, a mention might be a minor part of a broader discussion where the surrounding text focuses on other concepts, potentially injecting irrelevant context into the model's decision on the cross-document relation. We observe that definitions generally help mitigate these challenges by providing detailed explanations that highlight similarities or shared concepts that extend beyond the surface-level context. For example, \emph{conversational speech synthesis} and \emph{conversational speech processing tasks} (see Table \ref{tab:example_definitions}, row 3) where the first mention is part of a broader discussion on F0 control and speaking attitudes, and the second mention involves a focus on privacy-preserving features in speech processing. We note that this type of context can affect the quality of definitions as well; however, the retrieved global literature information may help mitigate problems in the local context, such that the definition generation model can often successfully generate a useful enough definition to assist the classification model.

\subsection{Impact of Relational Definitions} Following our exploration of how definitions generally improve model performance over non-definition approaches, we now turn to the impact of relational definitions compared to singleton definitions. We examine key cases where relational definitions outperform singleton definitions, particularly in capturing complex hierarchical or co-referential relationships, and discuss some trade-offs involved in both approaches.

Although singleton definitions often proved to be more effective than no definitions, we found they can often fall short (see also Table \ref{tab:example_definitions_errors} in the Appendix). In some cases, definitions share overlapping concepts without adequately distinguishing the unique roles of each mention.  Definitions may fail to capture the shared relationship between two mentions, especially when they focus on different aspects of the same underlying concept. This can lead the model to accentuate differences rather than recognize the commonality between the mentions. For instance (see also Table \ref{tab:example_relational_definitions}, row 1), the relational definitions more explicitly detail the relationship between \emph{word embedding matrix} and \emph{distributed representations of words}.
Similarly, in the case of \emph{simulated-annealing-based approach} and \emph{meta-heuristic approaches} (Table \ref{tab:example_relational_definitions}, row 2), the relational definitions highlight Simulated
Annealing as an approach in the broader umbrella of meta-heuristic approaches. In addition, relational definitions help clarify when there is no connection between two mentions in cases where singleton definitions do not provide sufficient discriminative information, e.g., in the case of \emph{neural network classifier} and \emph{network classification} (Table \ref{tab:example_relational_definitions}, row 3).

\paragraph{Relational definitions: Error cases} Relational definitions may sometimes emphasize similarities or overlaps irrelevant to the actual relation between mentions. For example, the relational definitions for \emph{Transformer model} and \emph{Equivariant Transformers (ETs)} (see Table \ref{tab:example_definitions_errors} in the Appendix) incorrectly identifies a hierarchical relationship, while the mentions refer to different architecture families.

\subsection{Retrieval definitions vs. GPT4o definitions}

Finally, we qualitatively examine how retrieval definitions differ to \texttt{GPT4o}-based definitions. 
We find that generally, retrieval generates more detailed and comprehensive definitions covering more diverse concept aspects, that allow detecting when one concept is more general than the other; conversely, \texttt{GPT4o} tends to generate more vanilla, less rich definitions (e.g., see Table \ref{tab:example_retrieval_definitions}, row 1 in the
Appendix). Additionally \texttt{GPT4o} definitions tend more often to fixate and over-rely on context, making the definitions focused on too narrow aspects of the context which do not allow to zoom out and detect abstractions and subsumptions (see example in Table \ref{tab:example_retrieval_definitions}, row 2 in the Appendix). These findings help explain why retrieval definitions appear to add several points in Hierarchy F1 vs. \texttt{GPT4o} definitions (Table \ref{tab:co-ref_hard_10} in the Appendix).

\definecolor{lightyellow}{HTML}{FFFFCC}
\definecolor{lightblue}{HTML}{ADD8E6} 

\section{Related Work}
\label{sec:related}
\paragraph{Coreference Resolution} Our task is closely related to the cross-document co-reference resolution task \cite{Yang2015AHD,Cattan2021CrossdocumentCR}. In scientific domains, CDCR presents unique challenges due to the abstract nature of technical concepts and the need to differentiate between closely related terms at varying levels of specificity \cite{cattanscico}. \citet{Ravenscroft2021CD2CRCR} further extend this line of work by introducing a cross-document, cross-domain coreference resolution task, which addresses coreference resolution between scientific articles and related news reports. Their work emphasizes the linguistic complexity introduced by domain-specific variations, highlighting significant limitations of the existing CDCR methods. \citet{Finin2009UsingWF} explored leveraging external structured knowledge through Wikitology, a knowledge base built from Wikipedia, DBpedia, and Freebase, demonstrating that incorporating structured semantic information substantially improves cross-document entity coreference performance. Recent advancements in coreference resolution have been reached with LLMs. As demonstrated by \citet{Le2023AreLL}, \citet{Gan2024AssessingTC}, LLMs like GPT-4, InstructGPT and Llama variants, when employed with prompt-based techniques, have shown strong performance in co-reference resolution tasks in the more popular within-document setting. \citet{yang-etal-2022-gpt} analyzed GPT models ability to resolve coreference using QA-based prompting. They found that while those models can generate reasonable answers, their coreference resolution performance is inconsistent and highly sensitive to prompt phrasing, which highlights some limitations of LLMs in those domains. In our approach, we explore LLMs in the challenging cross-document setting, in the scientific domain, and include also cross-document \emph{hierarchy} detection. We use both in-context learning and fine-tuning with dynamically generated (relational) definitions to improve CDCR performance.

\paragraph{Definitions in downstream NLP tasks}
Earlier works leveraged external knowledge to improve text understanding. \citet{AlHarbi2017LexicalDI} addressed lexical ambiguity in QA by incorporating WordNet-based definitions and domain-specific ontologies to refine word meanings. By integrating context knowledge, their approach improved word sense disambiguation, ensuring more precise retrieval in question answering. \citet{Yin2018TermDH} similarly used dictionary definitions to improve hypernymy detection, addressing the limitations of distributional embeddings, which infer meaning based on co-occurrence patterns in large corpora.  Although the usage of static definitions can be effective, they are often insufficient in our cross-document coreference and hierarchy task. Terms may have context dependent meanings that can vary across different papers, research fields, or even within the same document. In our approach, instead of relying on predefined dictionary definitions, we incorporate contextual definitions, generated with retrieved external knowledge in order to capture the nuanced semantic distinctions or similarities.

More recent works includes \citet{Munnangi2024OntheflyDA} who enhanced biomedical NER capabilities of LLMs in the few-shot setting by incorporating static definitions of canonicalized entities primarily from the UMLS knowledge base \citet{bodenreider2004unified}. In our method, we generate \emph{dynamic} singleton and relational definitions specifically tailored to the unique context of the mentions in scientific texts, by retrieving relevant literature. Our definitions are designed to enhance understanding of cross-document relationships. \citet{Xin2024LLMAELLL} recently proposed a method that uses LLMs as context augmenters to enhance entity linking (EL) with mention descriptions which are integrated into traditional EL models to improve linkage with canonical entities in general-domain KBs. \citet{Silva2020XTEET} introduce XTE (Explainable Text Entailment), an approach that explicitly leverages structured lexical definitions to improve semantic reasoning in RTE (Recognizing Textual Entailment). While their approach addresses entailment rather than cross-document coreference, it similarly demonstrates the benefit of augmenting neural models with structured concept definitions to enhance performance in relational reasoning tasks.

\paragraph{Relational Descriptions generation} There has been scarce work we are aware of on generating relational concept descriptions. \citet{Murthy2022ACCoRDAM} introduced a system named \textsc{ACCoRD} for generating diverse descriptions of how scientific concepts are related in order to provide varied perspectives on scientific concepts. The focus of \textsc{ACCoRD} was to build a small-scale user-facing system demonstration; unlike \textsc{ACCoRD}, our work uses relational definitions to directly improve NLP tasks such as CDCR and hierarchy inference. \citet{Brahman2020LearningTR} generate natural language definitions in the form of rationales to explain defeasible inference, a type of nonmonotonic reasoning where new information can strengthen or weaken prior conclusions. These definitions include both local rationales for individual concepts and relational explanations of how updates alter entailment. They are generated by prompting a pretrained language model with a premise, hypothesis, and update, and trained via distant supervision without retrieval. While focused on entailment rather than coreference, their work similarly highlights the value of textual definitions for modeling conceptual relationships.

\section{Conclusion}
\label{sec:conc}
We presented \projname, a novel approach based on dynamic definition generation and relational reasoning for enhancing the ability of Large Language Models (LLMs) to infer cross-document coreference and hierarchy relationships in scientific papers. Our method employs context-dependent and relational definitions generated through literature retrieval, along with a re-ranking based scheme to avoid combinatorial explosion across all potential mention pairs. \projname~has shown consistent improvements across both fine-tuning and in-context learning settings, with particularly strong gains on challenging subsets, and moderate improvements on the overall dataset. We showed that with our finetuning and dynamic definition integration approach, smaller open models can outperform much larger models. We also conducted in-depth analyses into the quality of generated definitions and failure modes which could be valuable to explore in future work. In addition, paired bootstrap tests provide further evidence that the improvements, particularly on the hardest subsets, are consistent across resampling.

Future research may focus on improving the generation and utilization of dynamic concept definitions, as well as improving the relational reasoning of LLMs when handling nuanced, fine-grained scientific concepts. Generating relational definitions can be computationally intensive at scale; we thus do not apply relational definitions across all possible mention pairs. Exploring ways to further optimize the efficiency of our re-ranking approach, or other approaches for explicit relational reasoning that avoid combinatorial explosion with infeasible costs, may lead to more scalable solutions for large-scale scientific applications. Another interesting direction is to investigate how the addition of structured metadata such as publication venue, field of study or year can help. For example, the term \emph{transformer} may be interpreted as a neural model in a machine learning conference but as power equipment in an electrical engineering journal. Metadata signals could enhance both the generation of contextual definitions and the prediction of cross-document relations, particularly in ambiguous or long-tail cases. In addition, while orthogonal to our contribution, we note that the \textsc{SciCo} dataset (which is over 3 times larger than the common CDCR benchmark ECB+ \cite{cybulska2014using}) could be expanded by collecting more annotations of coreference clusters and hierarchies across many papers from different domains, to better capture the breadth and scale of literature. While our approach improves performance, particularly on harder examples, it comes with additional computational cost due to relational definition generation and fine-tuning. This cost also carries environmental impact especially in large-scale settings. Developing more efficient alternatives for relational reasoning and knowledge integration is not only important for scalability but also for reducing environmental footprint. We leave exploration of lower-cost yet effective variants as a promising direction for future work. Finally, it would be exciting to consider further applications of our method, in more general cross-document tasks.



\section*{Acknowledgments}
We thank the reviewers and the action editor for their valuable feedback and suggestions. This research was supported by the Azrieli Foundation.

\bibliography{bib}
\bibliographystyle{acl_natbib}

\clearpage

\appendix
\section{Appendix}
\subsection{Definition Generation technical details}
\label{app:definition_generation}

\begin{figure*}
    \centering
    \includegraphics[width=\linewidth]{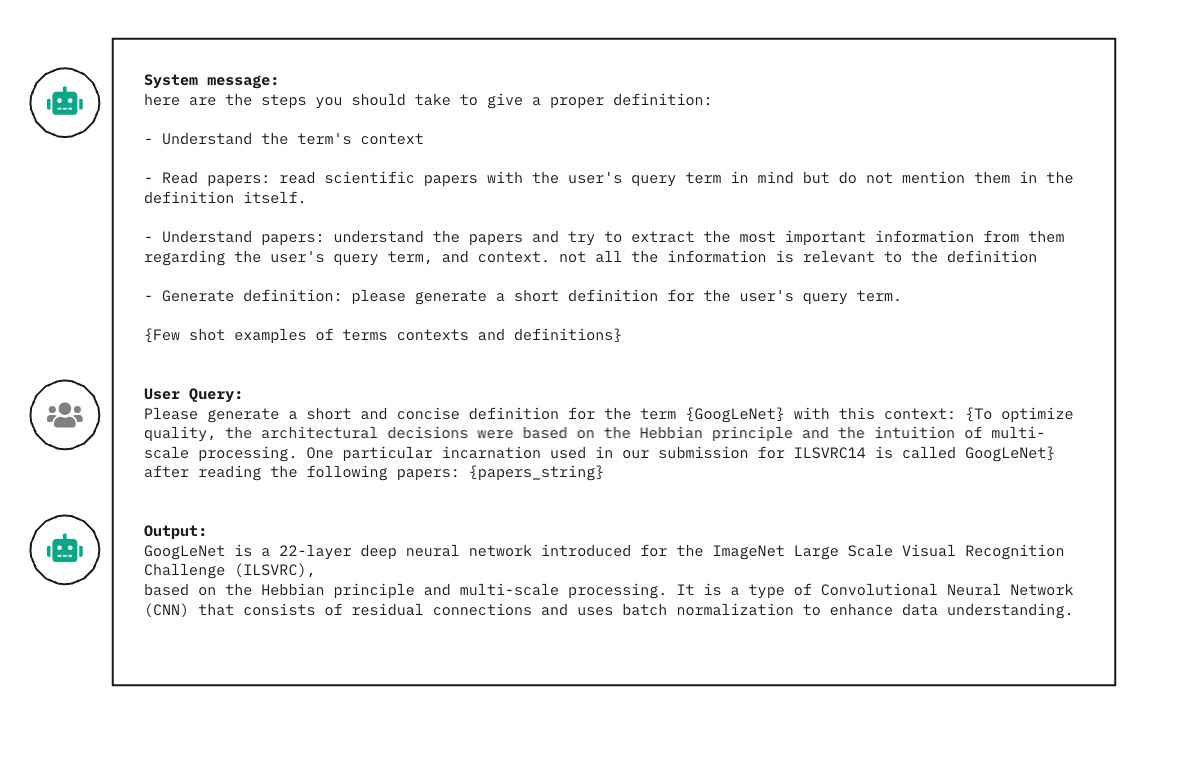}
    \caption{Singleton definition generation prompt.}
    \label{figure:singelton_def_generation_prompt}
\end{figure*}

\begin{figure*}
    \centering
    \includegraphics[width=\linewidth]{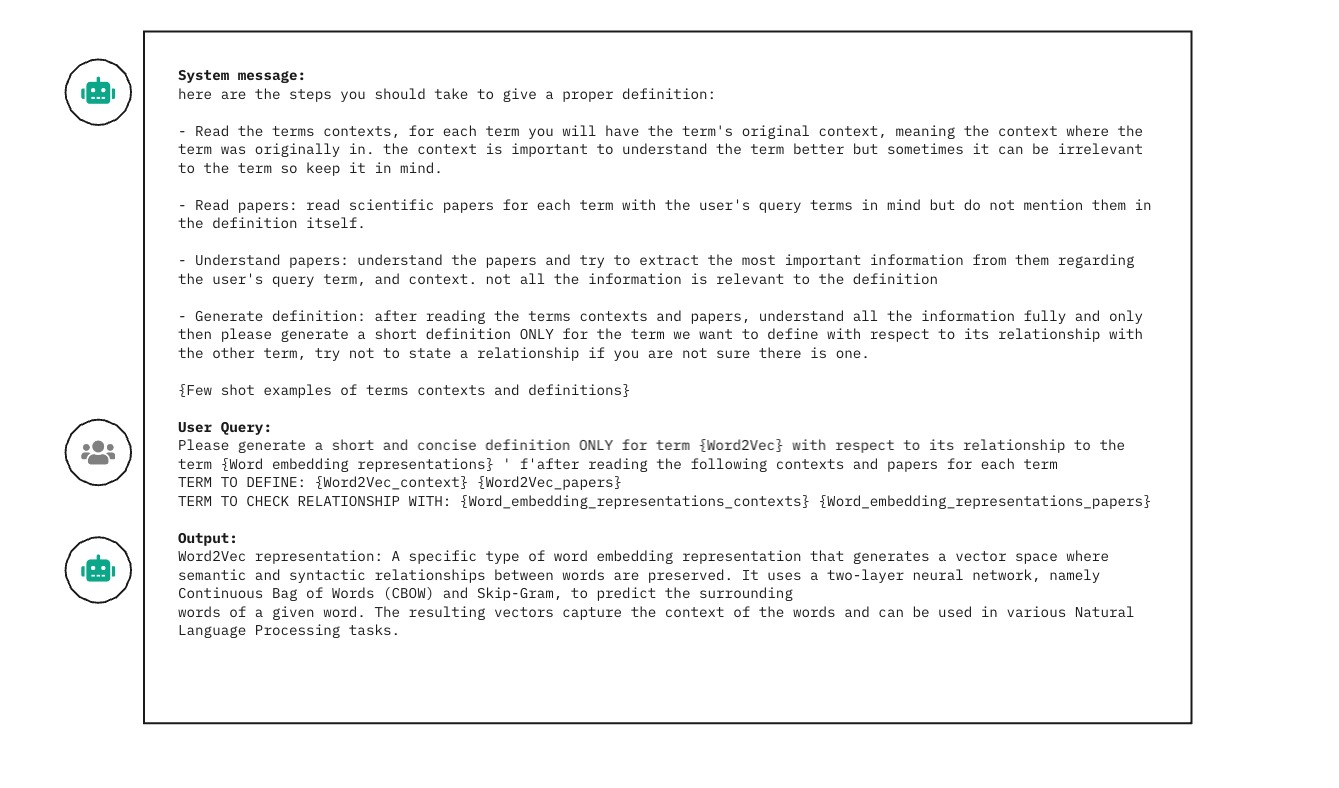}
    \caption{Relational definition generation prompt.}
    \label{figure:relational_def_generation_prompt}
\end{figure*}

For each definition generation iteration we retrieve the relevant papers, using \texttt{mxbai-embed-large-v1} \cite{emb2024mxbai,li2023angle} as an embedding model with state-of-the-art performance. 
We use the \texttt{Chroma DB}\footnote{ \url{https://github.com/chroma-core/chroma}} open-source vector database for fast retrieval using approximate nearest neighbor search. We rerank retrieved results with a cross-encoder (\texttt{mxbai-rerank-large-v1}\footnote{
\url{mixedbread.ai/docs/reranking/models}}. Results are then finally fed into the definition generation model, for which we use \texttt{Mixtral 8x7B}, a mixture of experts LLM \cite{Jiang2024MixtralOE}. We employ this method to produce singleton definitions for each individual mention and relational definitions (see Figures \ref{figure:singelton_def_generation_prompt}-\ref{figure:relational_def_generation_prompt} for specific prompts and \ref{app:hardware_lora} for hardware details and limitations).

\subsection{Additional Hardware and LoRA Hyper-Parameter Details}
\label{app:hardware_lora}

\paragraph{Hardware Setup and Runtime}
We employed 8 NVIDIA RTX A6000 48GB GPUs to generate both singleton and relational definitions through our RAG pipeline. Generating relational definitions only for the top-ranked mention pairs (§\ref{sec:method}) took roughly two weeks, while to generate relational
definitions for the entire test set, we would need roughly 40 days. For \textsc{Mistral-7B}, Fine-tuning was conducted on 3 NVIDIA RTX A6000 48GB GPUs over 2 epochs. The training was carried out with a batch size of 2, 4 gradient accumulation steps and a maximum sequence length of 1664 tokens (these choices were made due to our academic resources). We used a default learning rate of 2e-5 with a weight decay of 0.001. 

\paragraph{LoRA Configuration and Hyper-Parameters}
We use Low-Rank Adaptation of LLM (LoRA), a parameter efficient fine-tuning method \cite{Hu2021LoRALA} with the following parameters selected according to our resource constraints: lora\_alpha: 16, lora\_dropout: 0.1, lora\_r: 8, SEQ\_CLS as the task type and modules\_to\_save=["score"] in order to update and save the weights of the classification head we added to the model.

\subsection{ICL}
\label{app:dspy}

\paragraph{DSpy} We select 200 pairs for training and 60 pairs for development. To ensure balanced evaluation, we drew a random subset of examples, ensuring an equal number of labeled examples across the different relations. Within our DSPy program we incorporate chain-of-thought (CoT) reasoning and BootstrapFewShotWithRandomSearch optimizer.\footnote{\url{https://dspy-docs.vercel.app/docs/building-blocks/optimizers}} In CoT, a model is instructed to reason step by step to get to its answer, which can improve performance \cite{Wei2022ChainOT}. BootstrapFewShotWithRandomSearch is an enhanced version of the BootstrapFewShot teleprompter,\footnote{\url{https://dspy-docs.vercel.app/docs/deep-dive/teleprompter/bootstrap-fewshot}} which uses a teacher (GPT4 \citet{Achiam2023GPT4TR}) - student (GPT4o-mini) approach to generate few-shot demonstrations by sampling from the training data. During the training phase, the teacher model simulates potential outputs and rationales for labeled instances. The teacher model iteratively generates few-shot examples by attempting to predict the correct outputs based on the training data, learning from its own successes and refining also the generated rationale over time. This process is repeated multiple times, with a random search conducted over the generated demonstrations. The chosen program is the one that achieves the highest performance on the dev set. During each iteration on the test set, the program receives few-shot examples along with their bootstrapped rationales, assisting it in forming better rationales, thereby resulting in improved predictions.

\subsection{Additional Ablation study}
\label{app:ablation}

\paragraph{Relational definitions Ablation} To further examine the effect of relational definitions, we conducted an ablation study restricted to the \emph{Hard-10} subset (Table \ref{tab:relational_ablation}). Generating relational definitions for all mention pairs across the full SCICO test set is computationally infeasible, as it would require producing hundreds of thousands of pairwise definitions, therefore we focused on the smallest subset. Interestingly, while coreference scores improved only modestly, the hierarchy metrics saw larger gains, strengthening our hypothesis that relational definitions particularly help the model capture hierarchical structure. To balance feasibility and performance, we set the candidate selection threshold at 25\%, which we found to offer a practical trade-off between computational cost and accuracy.

\subsection{Statistical Significance Analysis}
\label{app:significance}

\paragraph{Statistical Significance} 
We evaluate whether improvements of our \textit{\textsc{\projname}\textsubscript{Relational}} model over the strongest baseline (\textit{\textsc{SciCo-Mistral-7B}}) are statistically reliable using one-tailed paired bootstrap tests over topics.

\paragraph{Method} 
For each subset (\emph{Hard-10}, \emph{Hard-20}) and metric, we compute the observed difference $\Delta$ and generate $20{,}000$ bootstrap samples by resampling topics with replacement. For each sample, we recompute the metric and record the resampled difference $\Delta_b$. The one-tailed $p$-value is the proportion of samples in which the relational model does not outperform the baseline.

\paragraph{Results} 
On the \emph{Hard-10} subset, improvements across all hierarchy metrics and CoNLL~F1 are statistically significant ($p < 0.05$). On the \emph{Hard-20} subset, the relational model yields statistically significant improvements in both Hierarchy~F1 ($p = 0.050$) and Hierarchy~F1-50\% ($p = 0.048$), while gains in Path Ratio and CoNLL~F1 remain positive but are not statistically significant. These patterns are consistent with our hypothesis that relational definitions most strongly benefit the most challenging cases.

\begin{table}[h]
\centering
\small
\begin{tabular}{l l r r}
\toprule
\textbf{Subset} & \textbf{Metric} & $\Delta$ & $p$ \\
\midrule
\emph{Hard-10} & Path Ratio      & +4.30 & 0.047 \\
\emph{Hard-10} & Hier F1-50\%    & +8.57 & 0.044 \\
\emph{Hard-10} & Hierarchy F1    & +6.72 & 0.047 \\
\emph{Hard-20} & Hierarchy F1    & +4.77 & 0.050 \\
\emph{Hard-20} & Hier F1-50\%    & +4.99 & 0.048 \\
\emph{Hard-20} & Path Ratio      & +0.48 & 0.113 \\
\bottomrule
\end{tabular}
\caption{
Paired bootstrap significance results (20k samples).
$\Delta$ denotes improvement of \textsc{Relational} over the \textsc{Mistral-7B} baseline on hierarchy metrics.
}
\end{table}

\begin{table}[h]
\centering
\small
\begin{tabular}{l l c c r}
\toprule
\textbf{Subset} & \textbf{Model} & F1 & $\Delta$ & $p$ \\
\midrule
\emph{Hard-10} & Relational   & 71.89 & +1.91 & 0.040 \\
\emph{Hard-20} & Relational   & 76.23 & +0.64 & 0.086 \\
\bottomrule
\end{tabular}
\caption{
Paired bootstrap significance results (20k samples) for CoNLL~F1.
Positive improvements on \emph{Hard-20} are not statistically significant.
}
\end{table}

\begin{table*}
\centering
\begin{tabular}{lcccc} 
\toprule
\textbf{Model} & \textbf{Coreference} & \multicolumn{2}{c}{\textbf{Hierarchy}} & \textbf{Path} \\
               & \textbf{CoNLL F1} & \textbf{F1} & \textbf{F1-50\%} & \textbf{Ratio} \\
\textit{\textsc{\projname}\textsubscript{Relational def 25\%}} & 71.90 & 56.98 & 45.93 & 53.85 \\
\textit{\textsc{\projname}\textsubscript{Relational def 60\%}} & 71.96 & 57.71 & 46.25 & 54.31 \\
\textit{\textsc{\projname}\textsubscript{Relational def 100\%}} & \textbf{72.33} & \textbf{58.15} & \textbf{46.82} & \textbf{55.17} \\
\bottomrule
\end{tabular}
\caption{Ablation for relational definitions on \emph{Hard-10} subset.}
\label{tab:relational_ablation}
\end{table*}

\begin{table*}[h!]
\small
\resizebox{0.99\textwidth}{!}{%
\begin{tabular}{@{}>{\raggedright\arraybackslash}p{5cm} p{10cm}@{}}
\toprule
\textbf{Context and Prediction} & \textbf{Definition: } \\ 
\midrule

\textbf{1}: ... in various tasks such as \textbf{\emph{semantic parser}}, question-answering ...  & \textbf{1:} A Natural Language Processing tool that maps natural language into ... enabling the interpretation and manipulation of meaning ... \\ \\

\textbf{2}: ... modeled \textbf{\emph{syntactic parsing}} and SRL jointly... 
\newline\textcolor{darkgreen}{\textbf{Ground Truth: $\centernot\Longleftrightarrow$}} 
\newline\textcolor{darkgreen}{\textbf{No Def Model: $\centernot\Longleftrightarrow$}} & \textbf{2:} The process of analyzing a sentence to determine its grammatical structure ... provide structural information about sentences, which can aid in understanding meaning ... 
\newline\textcolor{red}{\textbf{Singleton Prediction: $\Longleftrightarrow$}}\\

\midrule

\textbf{1}: ... Three adapted deep neural networks : \textbf{\emph{ResNet}}, LSTM ...  & \textbf{1:} A type of deep residual network, primarily used in CNNs, that consists of several stacked residual units. Each unit is a collection of convolutional layers, accompanied by a shortcut connection ... \\ \\

\textbf{2}: ... which are extracted by a \textbf{\emph{deep pre-trained ResNet}} ... 
\newline\textcolor{darkgreen}{\textbf{Ground Truth: $\Longleftrightarrow$}} 
\newline\textcolor{darkgreen}{\textbf{No Def Model: $\Longleftrightarrow$}} & \textbf{2:} A Residual Network that has been previously trained on a large dataset, allowing it to extract meaningful features from images. It is often used as a component in various computer vision tasks ... 
\newline\textcolor{red}{\textbf{Singleton Prediction: $\centernot\Longleftrightarrow$}}\\

\midrule

\textbf{1}: ... extracted from DCT coefficients to achieve \textbf{\emph{unsupervised segmentation of image objects}}  ...  & \textbf{1:} The process of dividing an image into distinct sections or clusters based on certain visual characteristics, without requiring any prior labeled data or manual intervention... \\ \\

\textbf{2}: ... we introduce group saliency to achieve superior \textbf{\emph{unsupervised salient object segmentation}} ... 
\newline\textcolor{darkgreen}{\textbf{Ground Truth: $\Longrightarrow$}} 
\newline\textcolor{darkgreen}{\textbf{No Def Model: $\Longrightarrow$}} & \textbf{2:} The process of identifying and separating the most prominent objects in an image or image collection... utilizes intrinsic features and relationships among images to locate and segment salient objects that maximize inter-image similarities and intra-image distinctness ... 
\newline\textcolor{red}{\textbf{Singleton Prediction: $\centernot\Longleftrightarrow$}}\\
\midrule
\textbf{1}: ... the inference speed is twice that of sparsemax in \textbf{\emph{Transformer model}} ...  & \textbf{1:} A type of artificial neural network that uses an encoder-decoder structure ... the concept of Equivariant Transformers extends the Transformer model ...\\ \\
\textbf{2}: ... We propose \textbf{\emph{Equivariant Transformers ( ETs )}}, a family of ...  
\newline\textcolor{darkgreen}{\textbf{Ground Truth: $\centernot\Longleftrightarrow$}} \newline\textcolor{red}{\textbf{No Def Model: $\Longrightarrow$}} & \textbf{2:} A type of Transformer model designed to incorporate prior knowledge on the transformation invariances of a domain ... 
\newline\textcolor{red}{\textbf{Relational Prediction: $\Longrightarrow$}}\\
\bottomrule
\end{tabular}
}
\caption{Examples of Definitions with wrong predictions. We show cases where definitions can negatively impact the model predictions. $\centernot\Longleftrightarrow$ - no relationship, $\Longleftrightarrow$ - co-reference, $\Longrightarrow$ - 1 $\rightarrow$ 2 parent-child relationship and $\Longleftarrow$ - 2 $\rightarrow$ 1 parent-child relationship.}
\label{tab:example_definitions_errors}
\end{table*} 

\begin{table*}[h!]
\resizebox{0.99\textwidth}{!}{%
\begin{tabular}{@{}>{\raggedright\arraybackslash}p{6cm} p{6cm} p{7cm}@{}}
\toprule
\textbf{Context and Prediction} & \textbf{Singleton Def} & \textbf{GPT4o Def} \\ 
\midrule

\textbf{1}: ... embodied gameplay on \textbf{\emph{representation learning}} in the context of perspective taking ... & \textbf{1}: A process in machine learning where machines automatically discover the representations needed for detection or classification from raw data. It involves developing algorithms to automatically learn and improve the feature representations ...  & \textbf{1}: The process by which artificial agents autonomously discover and encode useful information and features from their observations of an environment, enabling them to understand concepts such as occlusion, object permanence, free space, and containment without needing large labeled datasets for each new task. \\ \\

\textbf{2}: ... A line of research learn the \textbf{\emph{embeddings of graph nodes}} through matrix factorization ...
\newline\textcolor{darkgreen}{\textbf{Ground Truth: $\Longrightarrow$}} & \textbf{2}: the transformation of nodes in a graph into a set of vectors, where each vector represents a node and aims to capture the graph topology ... These embeddings are often used as inputs for other machine learning tasks such as ... \newline\textcolor{darkgreen}{\textbf{Correct Prediction: $\Longrightarrow$}} & \textbf{2}: the low-dimensional vector representations of nodes within a graph, learned through techniques like matrix factorization, which capture the structural and relational information of the nodes to facilitate various machine learning tasks ... \newline\textcolor{red}{\textbf{Wrong Prediction: $\centernot\Longleftrightarrow$}} \\

\midrule

\textbf{1}: ... we show how \textbf{\emph{context-based word embeddings}} , vector representations of words ... & \textbf{1:} A type of word representation NLP that captures the variations of word meanings based on their context ...  Context-based word embeddings differ from classic word embeddings as they provide word vector representations based on their context, thereby implicitly providing a model ... & \textbf{1}: Vector representations of words that capture their meanings based on the surrounding context, used to enhance search capabilities and aid in label prediction for incident reports. \\ \\

\textbf{2}: ... models outperforms these models even though they use pre - trained \textbf{\emph{word embeddings}} ... 
\newline\textcolor{darkgreen}{\textbf{Ground Truth: $\Longleftarrow$}} & \textbf{2:} A type of word representation that allows words with similar meaning to have a similar representation. They are often trained on large amounts of text data and can capture  ... \newline\textcolor{darkgreen}{\textbf{Correct Prediction: $\Longleftarrow$}} & \textbf{2}: The representation of words in a continuous vector space where words with similar meaning have similar vectors. \newline\textcolor{red}{\textbf{Wrong Prediction: $\centernot\Longleftrightarrow$}} \\ 

\bottomrule
\end{tabular}
}
\caption{Analysis of retrieval vs. \texttt{GPT-4o} definitions. Examples of \texttt{GPT-4o}-generated definitions that do not help accurately resolve relations, in contrast to retrieval-enhanced definitions. $\centernot\Longleftrightarrow$ - no relationship, $\Longleftrightarrow$ - co-reference, $\Longrightarrow$ - 1 $\rightarrow$ 2 parent-child relationship and $\Longleftarrow$ - 2 $\rightarrow$ 1 parent-child relationship.}
\label{tab:example_retrieval_definitions}
\end{table*}

\begin{table*}[!htbp]
\centering
\footnotesize
\renewcommand{\arraystretch}{1.15}
\begin{tabular}{@{}lccc@{}} 
\toprule
\textbf{Model} & \multicolumn{2}{c}{\textbf{Hierarchy}} & \textbf{Path Ratio} \\
               & \textbf{F1} & \textbf{F1-50\%} &  \\
\midrule
\textit{\textsc{\projname}\textsubscript{GPT-4o}} & 51.2 & 41.6 & 50.9 \\
\textit{\textsc{\projname} \textsubscript{RAG}} & 54.7 & 43.4 & 50.0 \\
\bottomrule
\end{tabular}
\caption{Ablation: Comparison between definitions generated by RAG and GPT-4o. It is noticeable that the RAG definitions provide an edge with the more challenging \emph{Hard-10} subset on the hierarchy scores.}
\label{tab:defablation_hard_10}
\end{table*}

\begin{table*}
\centering
\begin{tabular}{lcccc} 
\toprule
\textbf{Model} & \textbf{Coreference} & \multicolumn{2}{c}{\textbf{Hierarchy}} & \textbf{Path} \\
               & \textbf{CoNLL F1} & \textbf{F1} & \textbf{F1-50\%} & \textbf{Ratio} \\
\midrule
\textit{\textsc{SciCo-Longformer}} (Baseline) & 68.18 & 46.44 & 29.61 & 36.21 \\
\midrule
\textit{\textsc{SciCo-Mistral-7B}} & 75.59 & 54.59 & 44.53 & 54.79 \\
\textit{\textsc{\projname}\textsubscript{Singleton}} & 75.88 & 57.61 & 47.41 & 53.31 \\
\textit{\textsc{\projname}\textsubscript{Relational def}} & \textbf{76.23} & \textbf{59.36} & \textbf{49.52} & \textbf{55.27} \\
\bottomrule
\end{tabular}
\caption{Model results on the \emph{Hard-20} subset.}
\label{tab:combined_hard_20}
\end{table*}

\begin{table*}[h!]
\centering
\begin{tabular}{@{}lcccccc@{}}
\toprule
 & \multicolumn{3}{c}{Hierarchy} & \multicolumn{3}{c}{Hierarchy 50\%} \\
\cmidrule(lr){2-4} \cmidrule(lr){5-7}
 & Recall & Precision & F1 & Recall & Precision & F1 \\
\midrule
\textit{\textsc{SciCo-Longformer} (Baseline)} & 39.67 & 54.99 & 46.09 & 27.46 & 48.76 & 35.14 \\
\textit{\textsc{SciCo-Longformer}\textsubscript{Singleton def}} & 40.54 & 57.13 & 47.43 & 31.35 & 49.94 & 38.52 \\
\textit{\textsc{Phi-3-mini}} & 34.16 & 70.96 & 46.12 & 29.58 & 64.00 & 40.46 \\
\textit{\textsc{Phi-3-mini}\textsubscript{Singleton def}} & 36.01 & 69.75 & 47.50 & 31.12 & 63.01 & 41.66 \\
\textit{\textsc{SciCo-Mistral-7B}} & 47.62 & 70.49 & 56.84 & 42.32 & 60.22 & 49.70 \\
\textit{\textsc{\projname}\textsubscript{GPT-3.5 def}} & 44.35 & 70.30 & 54.39 & 39.46 & 62.79 & 48.46 \\
\textit{\textsc{\projname}\textsubscript{GPT-4o def}} & 46.77 & 74.00 & 57.31 & 41.98 & 65.33 & 51.12 \\
\textit{\textsc{\projname}\textsubscript{Singleton def}} & 49.76 & 67.61 & 57.33 & 43.86 & 59.74 & 50.58 \\
\textit{\textsc{\projname}\textsubscript{Relational def}} & 49.41 & 70.25 & 58.03 & 42.75 & 62.34 & 50.72 \\
\bottomrule
\end{tabular}
\caption{Recall, Precision and F1 according to the Cluster-level Hierarchy Score (§\ref{paragraph: evaluation_metrics}) for all models on the full test set.}
\label{tab:Hierarchy_all}
\end{table*}

\begin{table*}[h!]
\centering
\begin{tabular}{@{}lcccccc@{}}
\toprule
 & \multicolumn{3}{c}{Hierarchy} & \multicolumn{3}{c}{Hierarchy 50\%} \\
\cmidrule(lr){2-4} \cmidrule(lr){5-7}
 & Recall & Precision & F1 & Recall & Precision & F1 \\
\midrule
\textit{\textsc{SciCo-Longformer} (Baseline)} & 42.71 & 50.88 & 46.44 & 21.30 & 48.56 & 29.61 \\
\textit{\textsc{SciCo-Longformer}\textsubscript{Singleton def}} & 42.94 & 57.83 & 49.28 & 24.78 & 53.00 & 33.77 \\
\textit{\textsc{Phi-3-mini}} & 31.84 & 68.64 & 43.50 & 24.66 & 61.02 & 35.13 \\
\textit{\textsc{Phi-3-mini}\textsubscript{Singleton def}} & 36.86 & 63.68 & 46.79 & 26.44 & 60.61 & 36.83 \\
\textit{\textsc{SciCo-Mistral-7B}} & 44.73 & 70.02 & 54.59 & 35.09 & 60.93 & 44.53 \\
\textit{\textsc{\projname}\textsubscript{GPT-3.5 def}} & 40.81 & 68.81 & 51.23 & 32.62 & 62.38 & 42.84 \\
\textit{\textsc{\projname}\textsubscript{GPT-4o def}} & 44.50 & 71.60 & 55.30 & 35.43 & 63.13 & 45.39 \\
\textit{\textsc{\projname}\textsubscript{Singleton def}} & 53.34 & 67.33 & 57.61 & 39.35 & 59.63 & 47.41 \\
\textit{\textsc{\projname}\textsubscript{Relational def}} & 50.90 & 71.20 & 59.36 & 40.02 & 64.92 & 49.52 \\
\bottomrule
\end{tabular}
\caption{Recall, Precision and F1 according to the Cluster-level Hierarchy Score (§\ref{paragraph: evaluation_metrics}) for all models on the \emph{Hard-20} subset.}
\label{tab:Hierarchy_hard20}
\end{table*}

\begin{table*}[h!]
\centering
\begin{tabular}{@{}lcccccc@{}}
\toprule
 & \multicolumn{3}{c}{Hierarchy} & \multicolumn{3}{c}{Hierarchy 50\%} \\
\cmidrule(lr){2-4} \cmidrule(lr){5-7}
 & Recall & Precision & F1 & Recall & Precision & F1 \\
\midrule
\textit{\textsc{SciCo-Longformer} (Baseline)} & 39.96 & 52.73 & 45.46 & 20.49 & 44.55 & 28.07 \\
\textit{\textsc{SciCo-Longformer}\textsubscript{Singleton def}} & 41.78 & 51.69 & 46.21 & 23.53 & 39.61 & 29.52 \\
\textit{\textsc{Phi-3-mini}} & 26.57 & 64.46 & 37.63 & 18.26 & 56.02 & 27.54 \\
\textit{\textsc{Phi-3-mini}\textsubscript{Singleton def}} & 36.06 & 62.78 & 45.75 & 22.28 & 60.15 & 32.56 \\
\textit{\textsc{SciCo-Mistral-7B}} & 39.96 & 67.69 & 50.25 & 28.60 & 53.85 & 37.36 \\
\textit{\textsc{\projname}\textsubscript{GPT-3.5 def}} & 35.09 & 66.92 & 46.04 & 24.14 & 57.79 & 34.05 \\
\textit{\textsc{\projname}\textsubscript{GPT-4o def}} & 40.77 & 68.97 & 51.25 & 31.85 & 60.34 & 41.69 \\
\textit{\textsc{\projname}\textsubscript{Singleton def}} & 46.45 & 66.67 & 54.75 & 34.48 & 58.61 & 43.42 \\
\textit{\textsc{\projname}\textsubscript{Relational def}} & 48.07 & 69.93 & 56.98 & 36.51 & 61.89 & 45.93 \\
\bottomrule
\end{tabular}
\caption{Recall, Precision and F1 according to the Cluster-level Hierarchy Score (§\ref{paragraph: evaluation_metrics}) for all models on the \emph{Hard-10} subset.}
\label{tab:Hierarchy_hard10}
\end{table*}

\begin{table*}[h!]
\centering
\resizebox{\textwidth}{!}{
\begin{tabular}{@{}lccccccccccccc@{}}
\toprule
 & \multicolumn{3}{c}{MUC} & \multicolumn{3}{c}{B\textsuperscript{3}} & \multicolumn{3}{c}{CEAFe} & \multicolumn{3}{c}{LEA} & CoNLL \\
\cmidrule(lr){2-4} \cmidrule(lr){5-7} \cmidrule(lr){8-10} \cmidrule(lr){11-13} \cmidrule(lr){14-14}
 & R & P & F1 & R & P & F1 & R & P & F1 & R & P & F1 & F1 \\
\midrule
\textit{\textsc{SciCo-Longformer} (Baseline)} & 85.19 & 85.84 & 85.52 & 74.26 & 75.77 & 75.01 & 69.37 & 68.36 & 68.86 & 71.75 & 73.19 & 72.46 & 76.46 \\
\textit{\textsc{SciCo-Longformer}\textsubscript{Singleton def}} & 86.59 & 86.68 & 86.64 & 77.52 & 76.23 & 76.87 & 69.25 & 72.91 & 71.03 & 75.22 & 73.77 & 74.49 & 78.18 \\
\textit{\textsc{Phi-3-mini}} & 90.87 & 87.70 & 89.06 & 85.05 & 77.01 & 80.83 & 73.38 & 76.85 & 75.07 & 83.30 & 74.72 & 78.78 & 81.72 \\
\textit{\textsc{Phi-3-mini}\textsubscript{Singleton def}} & 90.99 & 88.06 & 89.50 & 85.09 & 78.35 & 81.58 & 73.86 & 77.88 & 75.81 & 83.36 & 76.17 & 79.60 & 82.30 \\
\textit{\textsc{SciCo-Mistral-7B}} & 92.51 & 86.71 & 89.51 & 86.72 & 75.05 & 80.47 & 73.03 & 78.11 & 75.49 & 85.19 & 72.77 & 78.49 & 81.82 \\
\textit{\textsc{\projname}\textsubscript{GPT-3.5 def}} & 91.42 & 87.17 & 89.24 & 85.22 & 75.91 & 80.30 & 72.94 & 77.31 & 75.07 & 83.50 & 73.67 & 78.28 & 81.54 \\
\textit{\textsc{\projname}\textsubscript{GPT-4o def}} & 91.91 & 86.63 & 89.19 & 86.07 & 74.22 & 80.10 & 71.68 & 77.08 & 75.58 & 84.40 & 71.84 & 77.62 & 81.62 \\
\textit{\textsc{\projname}\textsubscript{Singleton def}} & 90.46 & 88.28 & 89.36 & 83.39 & 78.09 & 80.66 & 74.11 & 77.62 & 75.82 & 81.58 & 75.85 & 78.61 & 81.95 \\
\textit{\textsc{\projname}\textsubscript{Relational def}} & 90.90 & 88.11 & 89.48 & 84.35 & 77.56 & 80.81 & 73.28 & 78.74 & 75.91 & 82.63 & 75.32 & 78.80 & 82.07 \\
\bottomrule
\end{tabular}
}
\caption{Coreference results for all the models on the entire SCICO test
set according to all coreference metrics: MUC, B\textsuperscript{3}
, CEAFe, LEA and CoNLL F1.}
\label{tab:co-ref_all}
\end{table*}

\begin{table*}[h!]
\centering
\resizebox{\textwidth}{!}{
\begin{tabular}{@{}lccccccccccccc@{}}
\toprule
 & \multicolumn{3}{c}{MUC} & \multicolumn{3}{c}{B\textsuperscript{3}} & \multicolumn{3}{c}{CEAFe} & \multicolumn{3}{c}{LEA} & CoNLL \\
\cmidrule(lr){2-4} \cmidrule(lr){5-7} \cmidrule(lr){8-10} \cmidrule(lr){11-13} \cmidrule(lr){14-14}
 & R & P & F1 & R & P & F1 & R & P & F1 & R & P & F1 & F1 \\
\midrule
\textit{\textsc{SciCo-Longformer} (Baseline)} & 78.38 & 86.40 & 82.19 & 58.04 & 75.71 & 65.71 & 60.03 & 53.61 & 56.64 & 55.08 & 73.26 & 62.88 & 68.18 \\
\textit{\textsc{SciCo-Longformer}\textsubscript{Singleton def}} & 81.95 & 85.49 & 83.69 & 66.37 & 73.20 & 69.62 & 60.12 & 59.86 & 59.99 & 63.46 & 70.54 & 66.81 & 71.10 \\
\textit{\textsc{Phi-3-mini}} & 86.28 & 87.71 & 86.99 & 76.11 & 75.23 & 75.67 & 64.56 & 64.28 & 64.42 & 73.80 & 72.70 & 73.24 & 75.69 \\
\textit{\textsc{Phi-3-mini}\textsubscript{Singleton def}} & 86.09 & 88.08 & 87.07 & 74.12 & 78.13 & 76.07 & 66.33 & 62.84 & 64.54 & 71.68 & 76.78 & 74.13 & 76.12 \\
\textit{\textsc{SciCo-Mistral-7B}} & 88.41 & 86.30 & 87.34 & 77.78 & 71.89 & 74.77 & 64.94 & 64.39 & 64.66 & 45.46 & 69.41 & 72.31 & 75.59 \\
\textit{\textsc{\projname}\textsubscript{GPT-3.5 def}} & 87.03 & 87.19 & 87.11 & 76.29 & 73.01 & 74.61 & 65.15 & 66.00 & 65.57 & 73.96 & 70.65 & 72.26 & 75.77 \\
\textit{\textsc{\projname}\textsubscript{GPT-4o def}} & 87.59 & 86.03 & 86.95 & 77.52 & 70.18 & 74.85 & 63.41 & 63.95 & 65.35 & 75.13 & 67.51 & 71.12 & 75.72 \\
\textit{\textsc{\projname}\textsubscript{Singleton def}} & 86.22 & 88.15 & 87.17 & 73.50 & 75.45 & 74.46 & 64.49 & 63.95 & 64.22 & 71.05 & 72.98 & 72.00 & 75.88 \\
\textit{\textsc{\projname}\textsubscript{Relational def}} & 86.59 & 88.19 & 87.39 & 74.59 & 76.65 & 75.71 & 63.74 & 65.25 & 64.45 & 72.24 & 72.98 & 72.61 & 76.23 \\
\bottomrule
\end{tabular}
}
\caption{Coreference results for all the models on the \emph{Hard-20} subset of the SCICO test
set according to all coreference metrics: MUC, B\textsuperscript{3}
, CEAFe, LEA and CoNLL F1.}
\label{tab:co-ref_hard_20}
\end{table*}

\begin{table*}[h!]
\centering
\resizebox{\textwidth}{!}{
\begin{tabular}{@{}lccccccccccccc@{}}
\toprule
 & \multicolumn{3}{c}{MUC} & \multicolumn{3}{c}{B\textsuperscript{3}} & \multicolumn{3}{c}{CEAFe} & \multicolumn{3}{c}{LEA} & CoNLL \\
\cmidrule(lr){2-4} \cmidrule(lr){5-7} \cmidrule(lr){8-10} \cmidrule(lr){11-13} \cmidrule(lr){14-14}
 & R & P & F1 & R & P & F1 & R & P & F1 & R & P & F1 & F1 \\
\midrule
\textit{\textsc{SciCo-Longformer} (Baseline)} & 76.37 & 85.15 & 80.52 & 53.85 & 71.84 & 61.55 & 52.79 & 49.24 & 50.96 & 50.85 & 68.91 & 58.52 & 64.34 \\
\textit{\textsc{SciCo-Longformer}\textsubscript{Singleton def}} & 78.33 & 85.35 & 81.69 & 57.15 & 72.27 & 63.83 & 51.22 & 49.44 & 50.32 & 54.25 & 69.20 & 60.82 & 65.28 \\
\textit{\textsc{Phi-3-mini}} & 82.25 & 86.78 & 84.45 & 68.65 & 72.20 & 70.38 & 56.38 & 56.38 & 56.38 & 65.92 & 69.15 & 67.50 & 70.40 \\
\textit{\textsc{Phi-3-mini}\textsubscript{Singleton def}} & 80.94 & 86.47 & 83.61 & 63.84 & 75.41 & 69.15 & 58.19 & 53.38 & 55.69 & 60.89 & 72.44 & 66.16 & 69.48 \\
\textit{\textsc{SciCo-Mistral-7B}} & 85.51 & 84.63 & 85.06 & 71.63 & 65.95 & 68.97 & 54.92 & 57.51 & 56.19 & 68.78 & 62.72 & 65.61 & 69.98 \\
\textit{\textsc{\projname}\textsubscript{GPT-3.5 def}} & 83.42 & 85.43 & 84.41 & 68.79 & 69.18 & 68.98 & 56.45 & 57.48 & 56.96 & 66.09 & 66.12 & 66.11 & 70.12 \\
\textit{\textsc{\projname}\textsubscript{GPT-4o def}} & 84.46 & 85.24 & 84.85 & 71.15 & 67.29 & 69.16 & 56.02 & 56.53 & 56.27 & 68.43 & 64.01 & 66.15 & 70.10 \\
\textit{\textsc{\projname}\textsubscript{Singleton def}} & 82.90 & 86.87 & 84.84 & 67.44 & 70.96 & 69.15 & 56.40 & 57.43 & 56.91 & 64.78 & 67.98 & 66.34 & 70.30 \\
\textit{\textsc{\projname}\textsubscript{Relational def}} & 85.90 & 84.47 & 85.18 & 71.92 & 68.52 & 70.18 & 60.86 & 59.78 & 60.31 & 69.08 & 65.56 & 67.28 & 71.89 \\

\bottomrule
\end{tabular}
}
\caption{Coreference results for all the models on the \emph{Hard-10} subset of the SCICO test
set according to all coreference metrics: MUC, B\textsuperscript{3}
, CEAFe, LEA and CoNLL F1.}
\label{tab:co-ref_hard_10}
\end{table*}











\noindent

\noindent

\noindent
\begin{figure*}
    \centering
    \includegraphics[width=\linewidth]{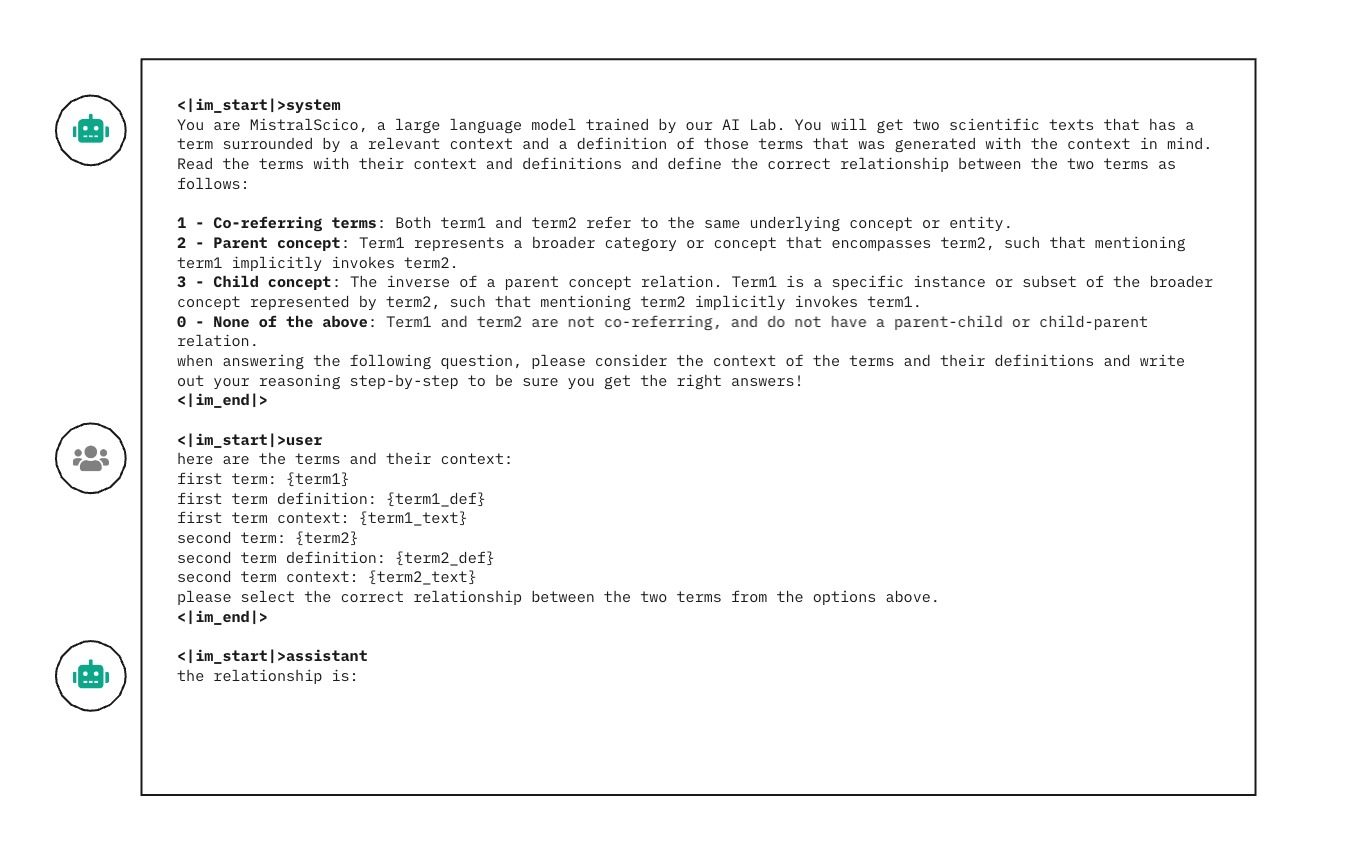}
    \caption{Chat ML finetuning prompt format.}
    \label{figure:finetuning_mistral_classification}
\end{figure*}

\noindent
\begin{figure*}
    \centering
    \includegraphics[width=\linewidth]{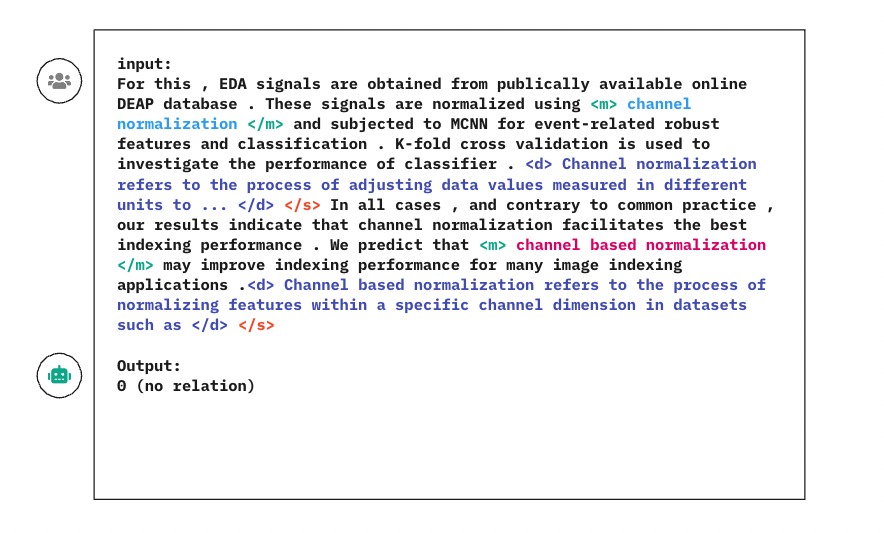}
    \caption{Longformer format.}
    \label{figure:Longformer_format}
\end{figure*}

\noindent
\begin{figure*}
    \centering
    \includegraphics[width=\linewidth]{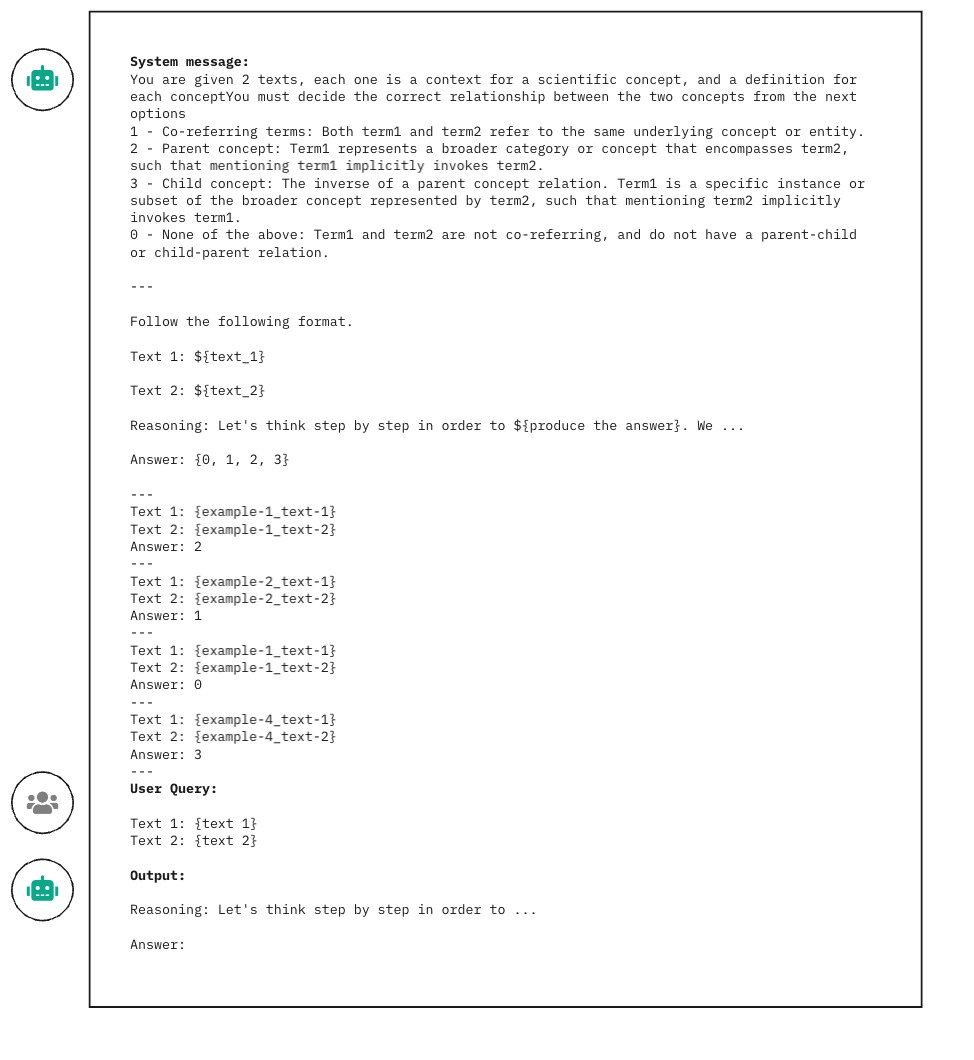}
    \caption{Few shots prompt format.}
    \label{figure:few_shots_no_def_format}
\end{figure*}

\noindent
\begin{figure*}
    \centering
    \includegraphics[width=\linewidth]{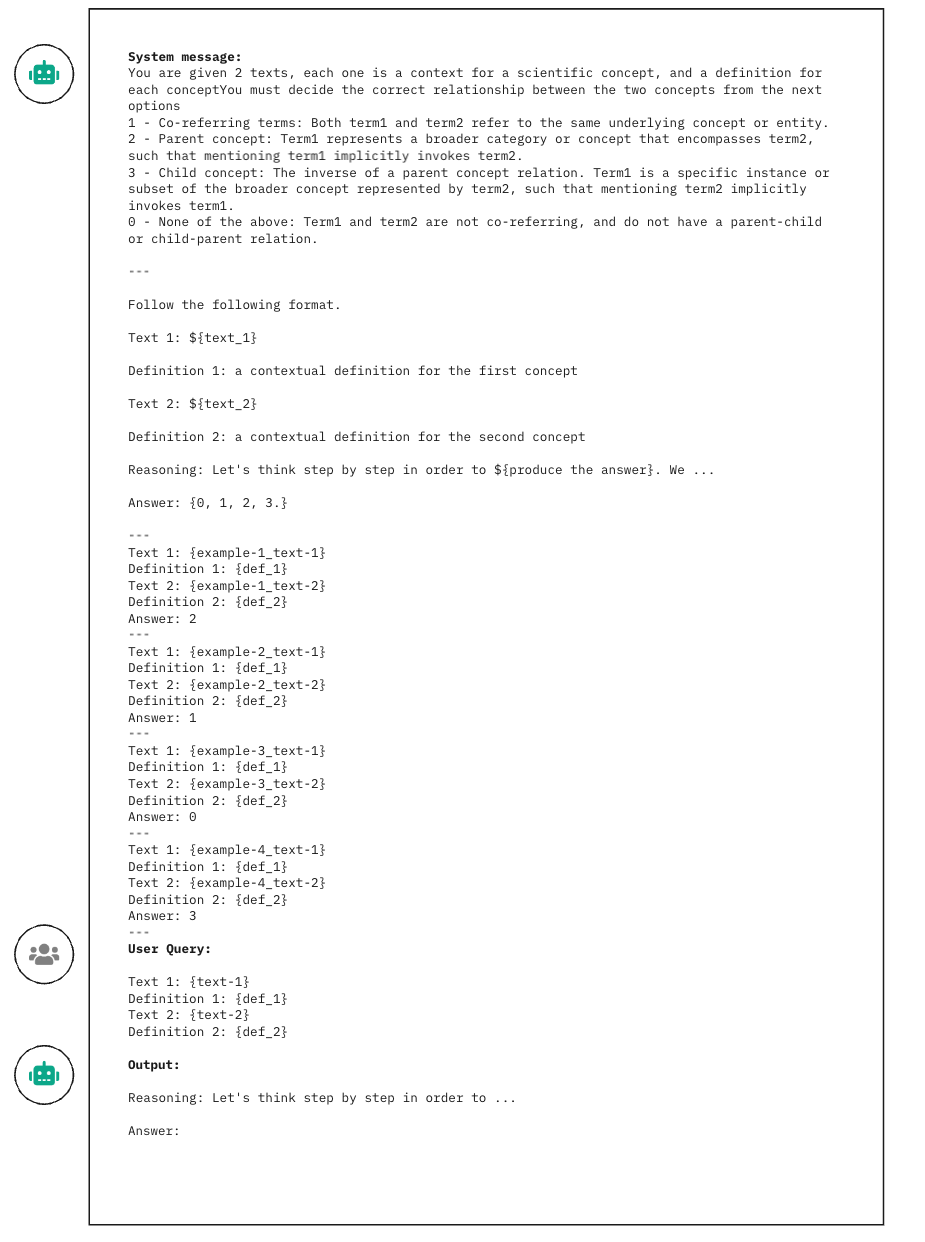}
    \caption{Few shots with definitions prompt format.}
    \label{figure:few_shots_with_def_format}
\end{figure*}

\end{document}